%% file: main.tex
\documentclass[preprint,11pt]{elsarticle}

\pdfoutput=1

\usepackage{amssymb}

\usepackage{algorithm}
\usepackage{algpseudocode}
\usepackage{color}
\usepackage{epsf}
\usepackage{psfrag}
\usepackage{epsfig}
\usepackage{graphicx}
\usepackage{amsfonts}
\usepackage{textcomp}
\usepackage{comment}
\usepackage{footnote}
\usepackage{tablefootnote}
\usepackage[flushleft]{threeparttable}
\usepackage{paralist}
\usepackage{mdwlist}
\usepackage{amssymb}
\usepackage{amsmath}
\usepackage{url}
\usepackage{verbatim}
\usepackage{alltt}
\usepackage{balance}
\usepackage{multirow}
\usepackage{epstopdf}
\usepackage{lipsum}% http://ctan.org/pkg/lipsum
\usepackage{float}
\usepackage{booktabs}
\usepackage{slashbox}
\usepackage[table]{xcolor}
\usepackage{array}
\usepackage{boldline}

\definecolor{green1}{RGB}{0,175,0}
\definecolor{blue1}{RGB}{0,150,200}
\definecolor{orange1}{RGB}{236,112,10}

\usepackage{caption}% http://ctan.org/pkg/caption
\usepackage[top=1.3in,bottom=1.3in,left=1.3in,right=1.3in]{geometry}
\journal{~}

\begin{document}

%\linenumbers  %% Delete in final version

\begin{frontmatter}

%% Title, authors and addresses

%% use the tnoteref command within \title for footnotes;
%% use the tnotetext command for theassociated footnote;
%% use the fnref command within \author or \address for footnotes;
%% use the fntext command for theassociated footnote;
%% use the corref command within \author for corresponding author footnotes;
%% use the cortext command for theassociated footnote;
%% use the ead command for the email address,
%% and the form \ead[url] for the home page:
%% \title{Title\tnoteref{label1}}
%% \tnotetext[label1]{}
%% \author{Name\corref{cor1}\fnref{label2}}
%% \ead{email address}
%% \ead[url]{home page}
%% \fntext[label2]{}
%% \cortext[cor1]{}
%% \address{Address\fnref{label3}}
%% \fntext[label3]{}

% \title{A Study on Writer Identification and Verification from Intra-variable Individual Handwriting}
\title{An Empirical Study on Writer Identification \& Verification from Intra-variable Individual Handwriting}

\author[label1,label2]{Chandranath~Adak}
  \ead{chandranath.adak@uts.edu.au}
  
\author[label3]{Bidyut~B.~Chaudhuri}
  \ead{bbcisical@gmail.com}
  
\author[label1]{Michael~Blumenstein}
  \ead{michael.blumenstein@uts.edu.au}
  
\address[label1]{Centre for AI, School of Software, University of Technology Sydney, Australia - 2007} 
\address[label2]{IIIS, School of ICT, Griffith University, Gold Coast, Australia - 4222}
\address[label3]{CVPR Unit, Indian Statistical Institute, Kolkata, India - 700108}

% \author{Chandranath Adak, Bidyut B. Chaudhuri, Michael Blumenstein}
% \address{Centre for AI, School of Software, University of Technology Sydney-2007, Australia\\
% IIIS, School of ICT, Griffith University, Gold Coast-4222, Australia\\
% CVPR Unit, Indian Statistical Institute, Kolkata-700108, India\\
% \{adak32, bbcisical\}@gmail.com, michael.blumenstein@uts.edu.au}

%% use optional labels to link authors explicitly to addresses:
%% \author[label1,label2]{}
%% \address[label1]{}
%% \address[label2]{}
%%%%%%%%%%%%%%%%%%%%%%%%%%%%%%%%%%%%%%%%%%%%%%%%%%%%%%%%%%%%%%%%%%%%%%%%%%%%%%%%%%%%%%%%%%%%%%%%%%%%%%%%%%%%%%%%%%%%%%%%%%%%%%
%  \author[label1]{Bidyut~B.~Chaudhuri\fnref{l1}}
%  \ead{bbcisical@gmail.com}
%  
%  \author[label1,label2]{Chandranath~Adak\fnref{l1}}
%  \ead{adak32@gmail.com}
%  
%  \fntext[l1]{Both the authors contributed equally to this work.}
%  
% \address[label1]{CVPR Unit, Indian Statistical Institute, India-700108}
% \address[label2]{School of ICT, Griffith University, Australia-4222}
%%%%%%%%%%%%%%%%%%%%%%%%%%%%%%%%%%%%%%%%%%%%%%%%%%%%%%%%%%%%%%%%%%%%%%%%%%%%%%%%%%%%%%%%%%%%%%%%%%%%%%%%%%%%%%%%%%%%%%%%%%%%%%

\iffalse
\author[label1]{Bidyut~B.~Chaudhuri}
\author[label1,label2]{Chandranath~Adak}

\address[label1]{CVPR Unit, Indian Statistical Institute, India-700108}
\address[label2]{School of ICT, Griffith University, Australia-4222}
\fi

%\ead{bbcisical@gmail.com, adak32@gmail.com}

\begin{abstract}
The handwriting of an individual may vary substantially with factors such as mood, time, space, writing speed, writing medium and tool, writing topic, etc. It becomes challenging to perform automated writer verification/identification on a particular set of handwritten patterns (e.g., speedy handwriting) of a person, especially when the system is trained using a different set of writing patterns (e.g., normal speed) of that same person. However, it would be interesting to experimentally analyze if there exists any implicit characteristic of individuality which is insensitive to high intra-variable handwriting. In this paper, we study some handcrafted features and auto-derived features extracted 
%strategically 
from intra-variable writing. Here, we work on writer identification/verification from offline Bengali handwriting of high intra-variability. To this end, we use various models mainly based on handcrafted features with SVM (Support Vector Machine) and features auto-derived by the convolutional network. For experimentation, we have generated two handwritten databases from two different sets of 100 writers and enlarged the dataset by a data-augmentation technique. We have obtained some interesting results.
\end{abstract}

\begin{keyword}
%% keywords here, in the form: keyword \sep keyword
%% PACS codes here, in the form: \PACS code \sep code
%% MSC codes here, in the form: \MSC code \sep code
%% or \MSC[2008] code \sep code (2000 is the default)
Intra-variable handwriting
\sep Writer identification  
\sep Writer verification. 
\end{keyword}

\end{frontmatter}

%% \linenumbers

%% main text
%\section{}
%\label{}

%\linenumbers

% \input{a}
\input{1intro}

\input{2related}
\input{3dataset}
\input{4preprocessing}
\input{5handcraftedFeature}

\input{6autoDerivedFeature}

\input{7writerIdentification}

\input{8writerVerification}

\input{9experiment}

\input{10conclusion}
%\input{tables}
% \input{figures}

%\input{reference} ==> bibliography substitute

%% The Appendices part is started with the command \appendix;
%% appendix sections are then done as normal sections
%% \appendix

%% \section{}
%% \label{}

%% If you have bibdatabase file and want bibtex to generate the
%% bibitems, please use
%%
\scriptsize
\section*{References}
\bibliographystyle{elsarticle-harv} 
%%  \bibliography{<your bibdatabase>}
\bibliography{Bibliography} % The references (bibliography) information are stored in the file named "Bibliography.bib"

%% else use the following coding to input the bibitems directly in the
%% TeX file.

%\begin{thebibliography}{00}

%% \bibitem{label}
%% Text of bibliographic item

% \bibitem{}
% 
% \end{thebibliography}
\end{document}

%% file: 1intro.tex
\section{Introduction}
\label{intro}

``Handwriting" is basically a kind of pattern. However, from the pre-historic era, it bears the connotation of human civilization. 
The handwriting instrument progressed from finger and wedge (on clay/sand and stone-based medium) to quill, pencil, fountain/ball-point pen (on parchment, papyrus/paper), and again finger (on the touch-screen of a smart device). Though the world is going fast towards a paperless e-world, ``handwriting remains just as vital to the enduring saga of civilization (\textendash Michael R. Sull)". 

For computer scientists, \emph{automated analysis of handwriting} is a recognized field of study owing to the ever-increasing complexity of extreme variations and having positive impacts on the fields of Forensics, Biometrics, Library Science and Data Science.

The handwriting pattern varies with person due to individual writing style. This may be termed as \emph{inter-class variance}. 
It is also noted that handwriting samples of a single person may vary extensively with various factors such as mood, time, space (geographical location), writing medium and tool. This is referred to as \emph{intra-class variance}. Sometimes, these inter-class and intra-class variations are termed as ``between-writer" and ``within-writer" variability, respectively \cite{13}. Even for excessive stroke variation among handwritten specimens of a particular writer, the writer and others having long exposure to his/her writing may still recognize it. Some implicit stroke characteristics may be the reason behind this ability.

In the field of forensics and biometrics, verifying/identifying a writer from a handwriting sample is sometimes essential (e.g., in the case of the ``2001 anthrax attacks"). Now-a-days, computer-assisted automated analysis is also quite popular in this application. Writer verification is a task used to authenticate a given document whether it is written by a certain individual or not. In writer identification, the goal is to match the writers to their handwriting specimens. The target of the writer identification/verification task is to maximize inter-variability and to minimize intra-variability.

\begin{figure}
  \centering
   \includegraphics[width=0.3\linewidth]{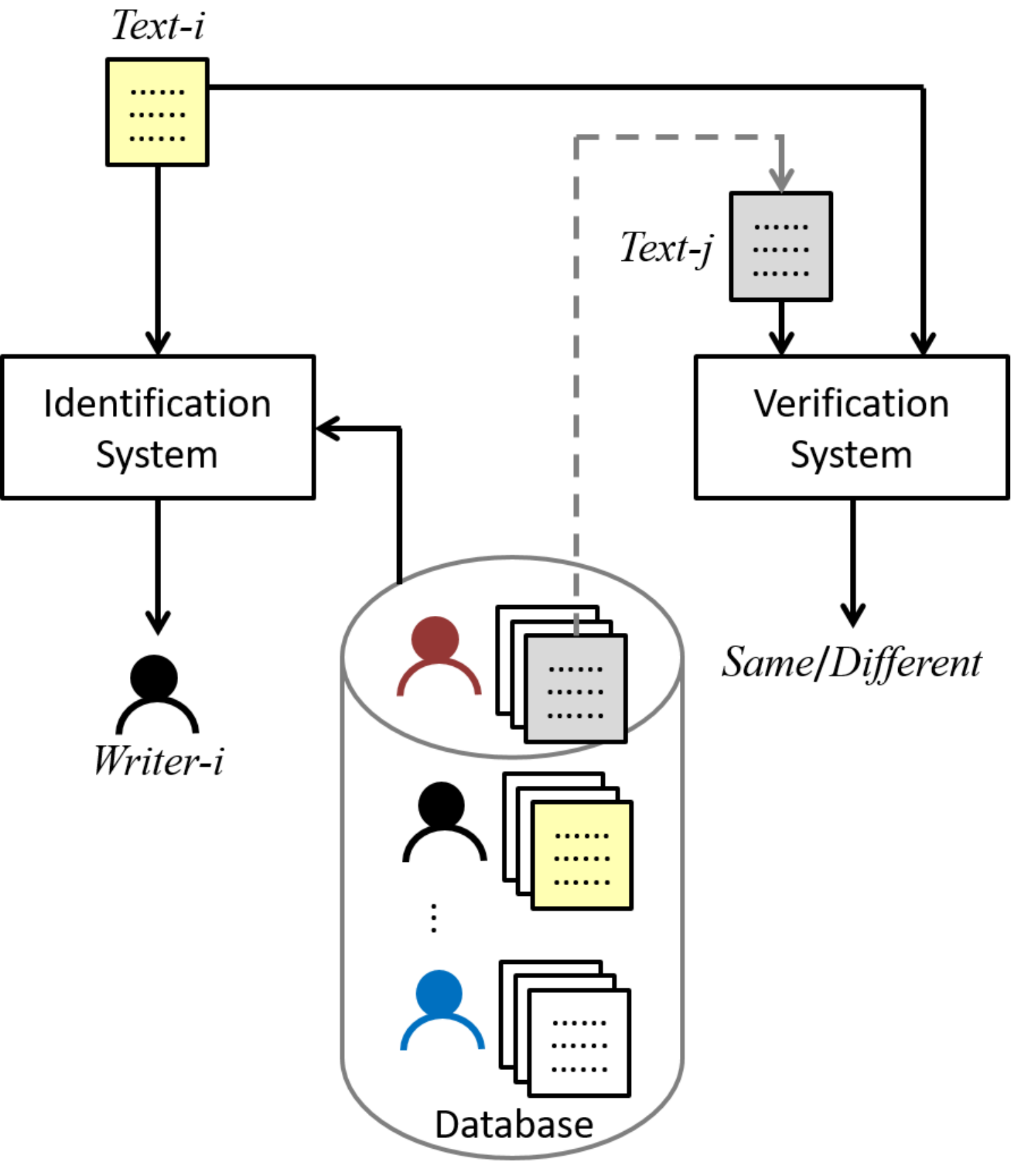}
  \caption{Ideal writer identification and verification system.}
  \label{fig:fig1}
 \end{figure}

The workflow of writer identification and verification approach is shown in Fig. \ref{fig:fig1}. Here, we have a database of handwritten texts with its known authors/writers. A query text sample \emph{Text-i} is input to the writer identification system to obtain its writer-id (\emph{Writer-i}) as an output with a certain degree of accuracy where the database provides support for the retrieval. In the writer verification system, two text samples \emph{Text-i} and \emph{Text-j} are fed to decide whether they are written by the ``\emph{Same}" or ``\emph{Different}" persons. The \emph{Text-i} is a query sample to be verified and the \emph{Text-j} may be fed from the database of known authors. 

In the document image analysis literature, interest has grown in the area of automated writer identification/verification for the last four decades. A detailed survey of the reported research works on this topic up to the year 1989 have been compiled in \cite{1}. Recent advancements on writer identification/verification can be found in \cite{2, 53}. 
Most of the past research work \cite{1, 2, 53} has focused on ideal handwriting generated in normal circumstances without paying much attention to the intra-variability. 

\begin{figure}
  \centering
   \tiny{(A1)}\fbox{\includegraphics[height= 1cm, width=0.43\linewidth]{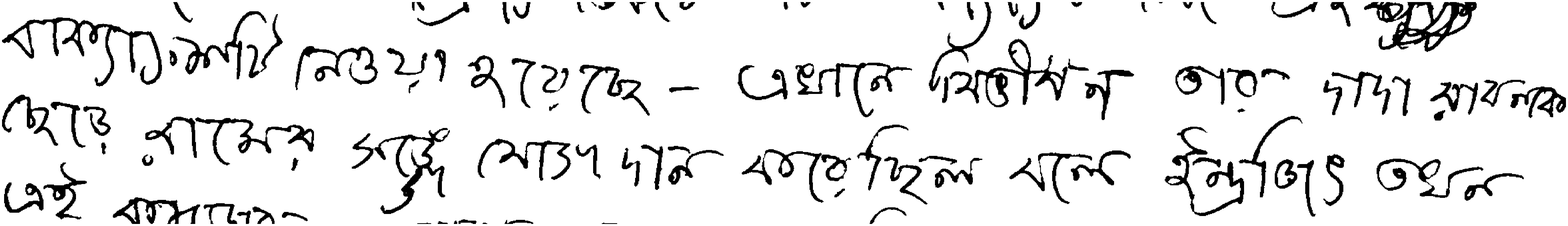}}
   ~(B1)\fbox{\includegraphics[height= 1cm, width=0.43\linewidth]{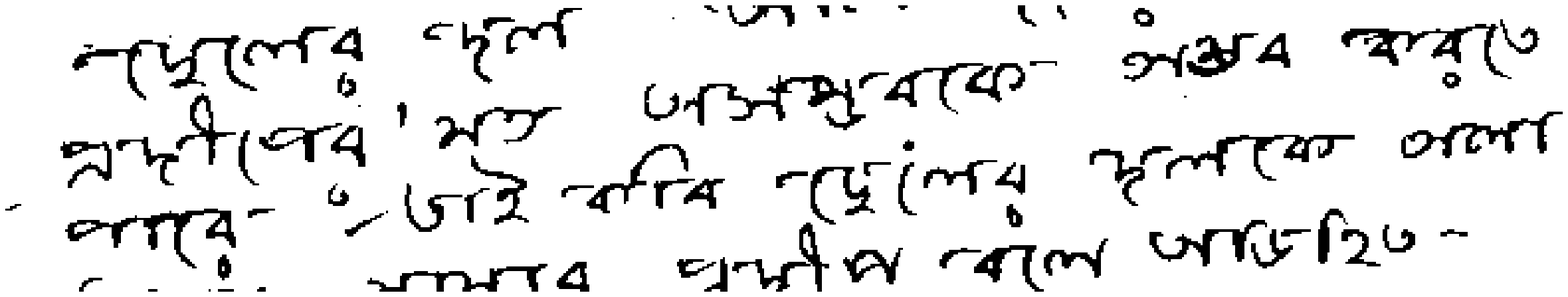}}\\[.1cm]
   (A2)\fbox{\includegraphics[height= 1cm, width=0.43\linewidth]{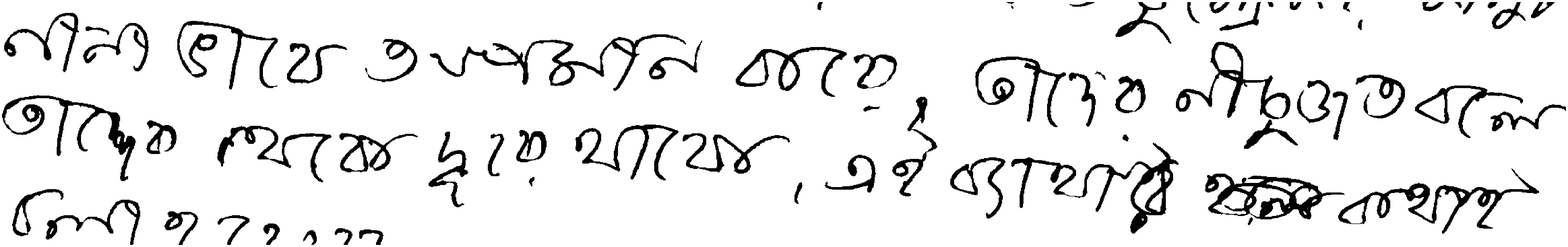}}
   ~(B2)\fbox{\includegraphics[height= 1cm, width=0.43\linewidth]{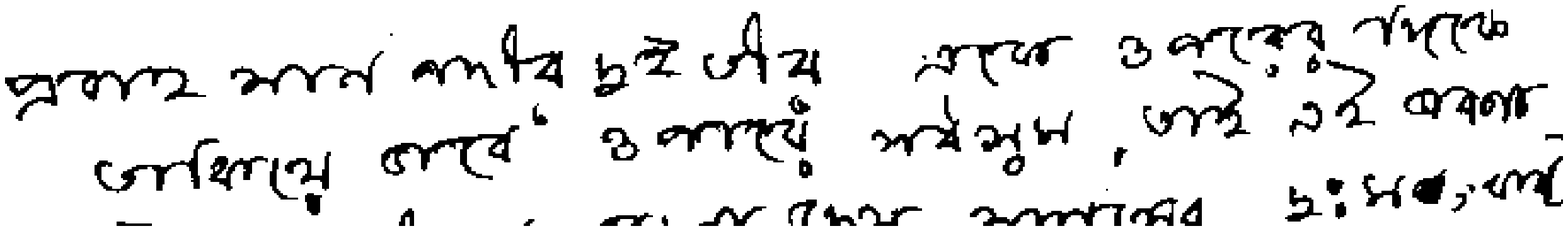}}\\[.1cm]
   (A3)\fbox{\includegraphics[height= 1cm, width=0.43\linewidth]{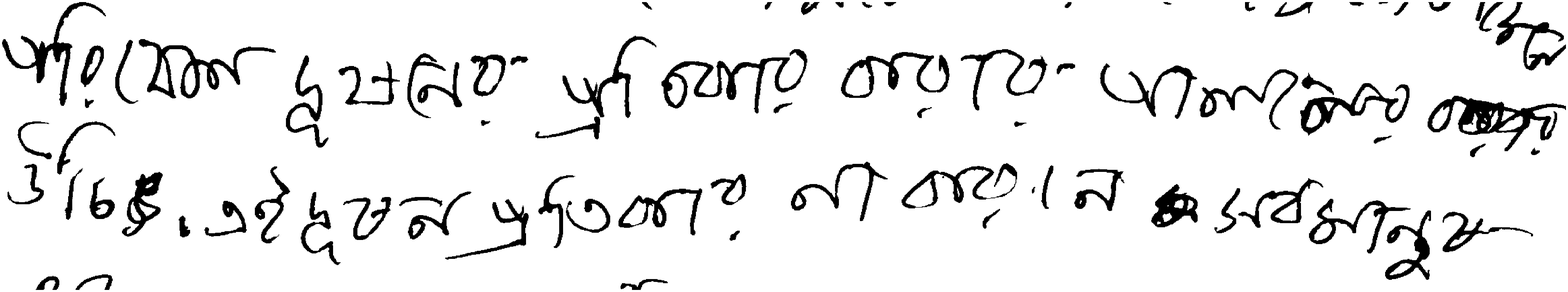}}
   ~(B3)\fbox{\includegraphics[height= 1cm, width=0.43\linewidth]{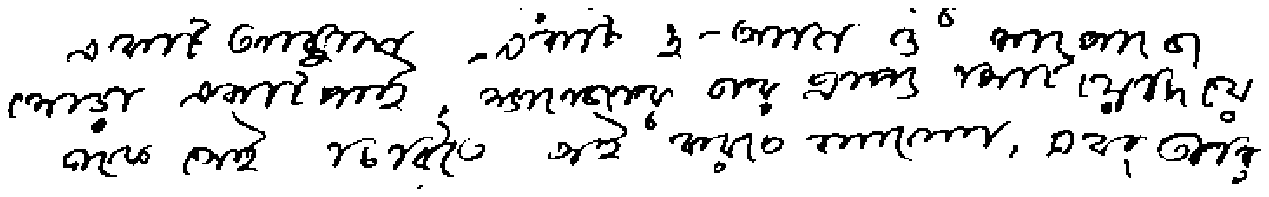}}
  \caption{Intra-variable Bengali handwritten samples, \emph{left column}: (A1), (A2), (A3) samples are of Writer-A, \emph{right column}: (B1), (B2), (B3) samples are of Writer-B. The intra-variability of Writer-A's samples is less compared to Writer-B's samples, as confirmed by handwriting experts.}
  \label{fig:fig2}
 \end{figure}

However, a situation may arise where an author needs to be verified based on a quickly written, unadorned handwritten manuscript, whereas only regular neat/clean handwriting with known authorship is available in the training database. Similar situations may occur where we need to identify the writer from unclaimed tidy handwriting, but the available database contains only careless untidy writing. 
In such a situation, the available writing of the person in the database and the test document written by the same individual can be highly dissimilar. In Fig. \ref{fig:fig2}, two sets of intra-varied handwritten sample of two individual writers are shown. 
Here, for example, we may need to verify whether the sample of Fig. \ref{fig:fig2}.(B1) is written by Writer-B, on the basis of B's handwriting of say Fig. \ref{fig:fig2}.(B3).

In this paper, we focus on the situation of intra-variation of individual handwriting to perform writer identification/verification. Ideally, within-writer variation should be less than the between-writer variation, which is the basis of the writer identification/verification task \cite{13}. However, where the intra-variation is relatively higher (refer to Fig. \ref{fig:fig2}), we need to find some handwriting features less sensitive to intra-variability and more sensitive to inter-variability.

We have generated two sets of database of intra-variable handwriting to deal with such realistic scenarios of writing identification/verification. Our database contains offline handwriting of \emph{Bengali} (endonym, \emph{Bangla}) script which is a fairly complex Indic script and used by more than 250 million people \cite{55, 5}. Recent advancements in writer identification on Indic scripts have been reported in \cite{6}. The general features of Bengali script can be found in \cite{5}. In connection with writer identification, some useful characteristics of Bengali handwriting are mentioned in \cite{6}, for example, matra/headline, delta, hole, coil shape, etc. We have noted that these Bengali handwriting characteristics along with classical handwriting characteristics (inter-text-line and inter-word gap, text-line skew, word/character slant, height/width of a character, text main-body height, character formation, etc.) usually vary with writing. 

In our generated database, the intra-variation is relatively higher than most of the existing databases in the literature \cite{53}, as confirmed by some handwriting experts. Therefore, here, we need to find some handwriting features which can decrease the intra-variability and increase the inter-variability.

%==========objectives===================
The \emph{applicability} of this work is as follows. 
\emph{First}, it will be helpful in the fields of forensics and biometrics for writer identification and verification. 
%
% \emph{Second}, this work studies the impact of working with missing data, 
% i.e., when a particular type of data is missing in the training set, how the system performs while testing with that type. 
\emph{Second}, this work studies the impact of working with absent data, 
i.e., when a particular type of individual writing is absent in the training set, how the system performs while testing with that type. 
\emph{Third}, this work may be useful in some applications of cultural heritage and library science. When some unpublished manuscripts of a scholar are found, the authorship is usually verified \cite{54}. Now, for the case of newly-discovered manuscripts, if they contain some unknown writing styles of the scholar, our work may provide some insights to analyze the authorship. 
\emph{Fourth}, this work may have a modest understanding of the progress of some diseases such as Parkinson's, Alzheimer's, Dysgraphia, Dyslexia, Tourette syndrome, etc., which affect handwriting. 
Here, before and after the disease progression, handwritten specimens have high intra-variability and provide some applicability of our work.

In this paper, we analyze the intra-variable handwriting for writer identification and verification tasks. The writer identification task is perceived as an $n$-class classification problem to classify handwritten documents in $n$ number of writer classes. On the other hand, writer verification is perceived as a binary classification, where a document is marked with ``Same" or ``Different" class, if it is written by the \emph{same} or \emph{different} writer of the given document, respectively. For these tasks, we extract two types of features, handcrafted and auto-derived \cite{52}, from an offline handwritten text sample set. Then the extracted features are classified to identify/verify a writer. 
In writer identification, SVM (\emph{Support Vector Machine}) is used for handcrafted features and some deep neural models are used for auto-derived features. 
In writer verification, we employ some similarity metrics on handcrafted and auto-derived features.

%================contribution======================
Our \emph{contribution} to this paper is the study of writer identification/verification on the intra-variable handwriting of an individual. Such a rigorous investigation is the earliest attempt of its type. 
For this study, we have generated two databases containing intra-variable handwriting in a controlled and uncontrolled way. 
The subgrouping of the uncontrolled database with respect to intra-variability is rather new. Here, the handwritten pages of the uncontrolled database are initially clustered with some deep features, and then finally grouped by confirmation of a classification technique which is pre-trained by the controlled database. 
We also propose two patch selection tactics to provide the input to the deep architectures without any normalization. 
Moreover, two writer identification strategies are introduced here, which are relatively new. 
The data augmentation technique is also a new addition here with respect to the offline handwritten data.

The rest of the paper is organized as follows. 
Section \ref{related}  discusses related work and Section \ref{3dataset} describes the experimental dataset generation procedure. 
Then Section \ref{4preprop} mentions the preprocessing step before entering into the methodology. 
After that Section \ref{5HCfeature} and Section \ref{6ADfeature} describe the handcrafted and auto-derived feature extraction techniques, respectively. 
The writer identification procedure is discussed in Section \ref{7WI}, followed by Section \ref{8WV} with a description of the writer verification process. 
The subsequent Section \ref{9Exp} is comprised of experimental results and discussions. 
Finally, Section \ref{10conclusion} concludes this paper.

%% file: 2related.tex
\section{Related Work}
\label{related}

With our both online and offline searching capacities, we have not found any direct work on intra-variable handwriting for writer identification/verification. In this section, at first, we cite some slightly related works mentioning intra-variability in handwriting. Then we briefly discuss some interesting work on writer investigation (i.e., identification/verification).  

\subsection{Intra-variability in Handwriting} 
The handwriting of a person may change for using various writing instruments. The handwriting alteration of a person using a pen and pencil has been studied in \cite{56}. Hilton \cite{57} reported that some pens can suppress the writing characteristics of an individual and may introduce intra-variability.

The handwriting of a person may change with the writing surface, more precisely, by the friction between the writing medium (paper) and tool (pen). Gerth et al. \cite{58} studied this within-writer variability when writing was performed on paper and a tablet computer.  

Handwriting also changes with time span and the age of a person \cite{14, 59}. Speedy writing in excitement, or writing with a stressed mind may degrade the writing quality, thus produces within-writer variability \cite{14, 60}.

Some diseases such as Parkinson's, Alzheimer's, etc. affect the handwriting of an individual \cite{59, 61}. Therefore, before and after such diseases, the writing shows intra-variability.  

Alcohol consumption also changes the individual's writing style and provides an example of intra-variable handwriting \cite{25}.

Such \emph{mechanical} (e.g., writing instrument, surface, etc.), \emph{physical} (e.g., illness, aging, etc.), \emph{psychological} (e.g., excitement, anger, mood, etc.) factors may cause large individual handwriting variation \cite{14}.

\subsection{Writer Investigation}
The root of writer analysis can be found around 1000 B.C. when a few Japanese scholars studied the bar formation in writing to judge personal characteristics \cite{62}. However, with the boom of automation, during the late 20\textsuperscript{th} century, such investigations had also started to be automated. In the beginning of the 21\textsuperscript{st} century, the ``9/11 attacks'' and ``2001 anthrax attacks'' have escalated automated writer identification/verification research.

In the document image analysis domain, automated writer investigation research is being performed during the last four decades. Plamondon and Lorette \cite{1} surveyed major works in this field up to the year 1989. After that, the offline writer investigation research up to the year 2007, is described in \cite{13}. The recent advancements in this field can be found in \cite{53}. The writer investigation research work on Indic scripts is discussed in \cite{6}. However, most of the past research has dealt with ideal handwriting without degradation \cite{1, 13, 53}. But, a handwritten page may contain various artifacts such as struck-out/crossed-out writing \cite{4, 40, 63}, doodles \cite{8}, ruled lines, printed text, logos, stamps, etc. \cite{3}. 

Chen et al. \cite{3} studied the impact of ruled line removal on writer identification. They \cite{3} showed that the performance improved by retaining, instead of deleting the ruled lines. 
Another work in \cite{4} reported the effect of struck-out texts on writer identification. The authors \cite{4} experimentally showed that the presence of struck-out texts in a handwritten document degrades the writer identification performance. 
A preliminary work on crossed-out text removal with its effect on writer identification is presented in \cite{63}.

Instead of handwritten text, sometimes writer inspection has been performed on unconventional notations such as musical scores \cite{64}, sketches \cite{65}, etc.

In the literature, writer identification has been tried with training on one script and testing on another \cite{66, 67}. In \cite{66}, English and Greek scripts were used, while in \cite{67}, English and Bengali scripts were reported.

%% file: 3dataset.tex
\section{Experimental Dataset}
\label{3dataset}

For our experimental analysis, we needed a database of intra-variable handwritten samples of each writer. We did not find any such publicly available database. Therefore, we had to generate a new database containing such handwriting specimens. We made two offline databases on the basis of a handwriting collection strategy, namely \emph{controlled} and \emph{uncontrolled} databases.

\subsection{Controlled Database ($D_c$)}
\label{sec.3.1.Dc}

In this case, all volunteers chosen for supplying data were aware of the modality of our experiments. For controlled database collection, we focused on intra-variability occurring mostly due to writing speed \cite{29}. 
It is noted that typically speedy writing is more distorted than the usual handwriting of a writer, and this inference was confirmed by some handwriting experts.

Collecting previously written samples at various speeds from many writers is quite challenging, since no one typically monitors and records his/her writing speed unless required for a specific reason. Therefore, we collected handwriting under a controlled setup, as follows. 

All volunteers were instructed to write at various speeds. At first, they were advised to write at their normal speed of writing. Then they were instructed to write faster than their normal writing speed. Finally, they were requested to write at a slower speed than normal. We labelled these as \emph{medium} (or, normal), \emph{fast}, and \emph{slow} sets of handwriting samples, respectively.

At the stage of full-page writing, we noted the total time $(t)$ of writing using a stop-watch and computed the total length $(l)$ of writing strokes on a page. The handwriting speed $(s)$ in a full page is calculated as stroke length per unit time, i.e., $s=l/t$. The stroke length is computed from counting the number of object pixels from the thinned version of the writing strokes. For thinning purpose, the \emph{Zhang-Suen thinning} method \cite{27} was used which worked better than some other techniques \cite{24}. Additionally, in the thinned version, a spurious branch having length less than half of the average stroke-width is pruned.

The reaction time ($\approx$ 200 millisecond) \cite{26} for using a stop-watch is negligible with respect to the objective of our task. We prepared our experimental setup to write on a white 70 GSM ($g/m^2$) A4 page and placing it on a horizontally plane surface with a smooth ball-point pen having black/blue ink. We created our database in offline mode, and did not use any digital pen/surfaces which could obstruct the individual writing habits. 
Here, our primary aim is to capture the intra-variability of handwriting, on which our technique can work adequately without using accurate, ultramodern speed measuring instruments.

For a writer $W_i$, a fast handwritten page is chosen where handwriting speed ($s_i$, a scalar quantity) is greater than a threshold $T1_i$, i.e., $s_i>T1_i$. 
For slow handwriting, $s_i<T2_i$, where $T2_i$ is another threshold. 
The medium handwritten page is chosen where $T1_i \ge s_i \ge T2_i$. The medium or normal handwriting speed is captured at first from multiple handwritten pages of a writer $W_i$. 
% We calculate the average medium speed ($\mu_{si}$) from these pages and also compute the standard deviation ($\sigma_{si}$) of the medium speed. 
We calculate mean ($\mu_{si}$) and standard deviation ($\sigma_{si}$) of medium speed from these pages.
Here, we use $T1_i=\lceil \mu_{si} + \alpha_s.\sigma_{si}\rceil$ and $T2_i=\lceil \mu_{si} - \alpha_s.\sigma_{si}\rceil$. We set $\alpha_s = 2$, which is decided empirically.

Each writer wrote multiple pages which were ordered in terms of the writing speed. We chose top speedy two pages from the \emph{fast} handwritten samples, two lowest-speeded pages from the \emph{slow} writing, and middle-speeded two pages from the \emph{medium} writing.

Although many volunteers contributed for our database generation, we chose 100 writers whose handwriting patterns varied structurally due to writing speeds, as advised by some handwriting experts. This is performed in order to generate a database of intra-variable handwriting.

The writers were native Bengali from West Bengal, India. The finally selected 100 writers were in the age group of 12-42 years having academic backgrounds from secondary school to university level. The ratio of male to female writers in this database is 14:11.

This controlled dataset ($D_c$) contains 600 pages where each writer contributed 6 pages of Bengali handwriting. Among 6 pages, 2 pages are of \emph{fast} handwriting speed, 2 pages are of \emph{medium} and the remaining 2 pages are of \emph{slow} speed. In other words, we have 3 sets of handwriting namely $S_f$ (\emph{fast}), $S_m$ (\emph{medium}), $S_s$ (\emph{slow}), each containing 2 handwritten pages of every 100 writers.

\subsection{Uncontrolled Database ($D_{uc}$)}
\label{sec.3.2.Duc}

For generation of the uncontrolled database, the writers should be unaware of our experiments before the data collection. However, people generally perform their daily writing at a normal pace using various pens and papers, and a uniform data collection setup is missing. Here, we came up with a different strategy for this type of database generation.

After discussions with some handwriting experts, we note that in real-life school examinations, the students generally write at various speeds, since the examination time is limited and the usual target is to score good marks by answering all questions within that stipulated period. Therefore, the students/writers are in a hurry when the clock is ticking towards the end of exam. However, here some behavioral/psychological aspects \cite{28, 29} may influence intra-variable handwriting, besides the writing-speed, due to anxiety, nervousness to finish, panic of a low score, stress to recollect the answer, etc.

Therefore, we have selected a real school-exam scenario to collect suitable data by maintaining uniformity for this data collection. Some details of such situation are as follows:

%\begin{enumerate}
 {i)}\emph{Time}: The examination duration was fixed as three hours. Here, the time works as a constraint to ascertain the increase/decrease of writing speed and individual variability.
 
 {ii)}\emph{Question type}: We selected a 100-mark Bengali literature examination paper, where most of the question types were broad subjective to be answered in many sentences.

 {iii)}\emph{Script}: The answers were to be written in Bengali script.

 {iv)}\emph{Paper}: For writing the answers, white pages of 70 GSM ($g/m^2$) with a fixed size of 215.9 $\times$ 355.6 $mm^2$ were provided.

 {v)}\emph{Pen}: The writers used their own pens. Most of the pens were of black/blue ink with 0.5 - 1.0 $mm$ ball-point tip.

 {vi)}\emph{Writer}: The writers were native Bengali from West Bengal, India, and students of VIII - XII grade Bengali-medium public schools. Their age ranged between 13 - 19 years. All these writers were different from the volunteers participated in controlled database generation.
%\end{enumerate}

The exam marking was performed by school teachers. For our task, we chose the answer script of a student who scored at least 40\% and wrote at least 6 full pages. A total of 153 writers were chosen in this way. Among these writers, 67 persons wrote 6 pages each, 11 persons wrote 7 pages each, 54 persons 8 pages each, 21 persons 9 pages each. Therefore, 153 writers contributed a total of 1100 pages.

Now, we have 1100 unlabelled pages of 153 writers. We label each page in one of the 3 groups of intra-variable writings, say, $S'_{fa}$, $S'_{ma}$ and $S'_{sa}$. For this grouping, at first, we use auto-derived feature-based clustering techniques, as follows.

We use the front part of GoogLeNet \cite{31} for feature extraction, since this architecture provides encouraging accuracy with comparatively lesser dimension of feature vector \cite{20}. From each page, an $n_p$ number of text-patches of size 224 $\times$ 224 is chosen arbitrarily. This $n_p$ is set empirically to 400. For each patch, we have obtained 1024-dimensional deep feature vector from the \emph{avg pool} layer of GoogLeNet \cite{31}. For clustering, we choose various clustering algorithms such as K-means, Fuzzy C-means, Minibatch K-means, Expectation-Maximization with Gaussian-Mixture-Models (EM with GMM), and Agglomerative Hierarchical Clustering (Agglo\_Hierarchical) \cite{22}.

Let us assume that p$_{ij}$ denotes patch\textsubscript{$i~|~i=1,2,\ldots,400$} of page\textsubscript{$j~|~j=1,2,\ldots,1100$}. Actually, p$_{ij}$ represents a 1024-dimensional feature vector obtained from patch\textsubscript{$i$} of page\textsubscript{$j$}. For p$_{1j}$, we obtain a cluster plot CL$_1$ by patch$_1$'s of all the 1100 pages. Similarly, cluster plot CL$_2$ is obtained from patch$_2$'s of all 1100 pages. And so on, patch$_{400}$'s of all 1100 pages produce cluster plot CL$_{400}$. 
These 400 cluster plots are ordered with a corresponding patch; i.e., patch\textsubscript{i}'s (or, the $i^{th}$ patch) of all pages produce CL$_i$. 
% Moreover, we obtain 600 (fixed empirically) unordered cluster plots, chosen arbitrarily. 
Moreover, we obtain 600 (fixed empirically) cluster plots unordered with patches which are chosen arbitrarily. 
Therefore, we now have 1000 ($=400+600$) cluster plots. 
Each cluster plot contains 1100 patches (i.e., patch-based feature points), where each patch represents each of the 1100 pages. 
From these 1000 cluster plots, using the \emph{majority rule}, we label each page into three groups/clusters $S'_{fa}$, $S'_{ma}$ and $S'_{sa}$. For example, suppose in all the 1000 cluster plots, 
majority %maximum number 
of the patches of page$_1$ (p$_{i1}$) fall in the cluster $S'_{fa}$; then the page$_1$ is put into the group $S'_{fa}$.

To find which clustering algorithm among K-means, Fuzzy C-means, Minibatch K-means, EM with GMM, and Agglo\_Hierarchical, would work well for intra-variable handwriting, we use an external evaluation criterion, called NMI (\emph{Normalized Mutual Information}) score \cite{68}. For this, we perform a similar feature extraction strategy and clustering techniques on the controlled dataset $D_c$, which contains the ground-truth. 
Employing various clustering techniques on $D_c$, we have obtained the NMI scores, as presented in Table \ref{tab:tablex}. The Agglo\_Hierarchical method worked well on $D_c$, so we use this clustering technique here also on uncontrolled data.

\begin{table}%[!htb]
%\scriptsize
\tiny
\caption{Clustering method evaluation on $D_c$}
\centering
\begin{tabular}{c|c}
\hline
{\bf{Clustering method}} & {\bf{NMI score}}\\ \hline %\hline
K-means & 0.637\\ \hline 
Minibatch K-means & 0.610\\ \hline 
Fuzzy C-means & 0.646\\ \hline 
EM with GMM & 0.679\\ \hline 
Agglo\_Hierarchical & {\bf{0.684}}\\ \hline 
\end{tabular}\label{tab:tablex}
\end{table}

Up to this point, the 1100 unlabelled pages of 153 writers are clustered into 3 groups ($S'_{fa}$, $S'_{ma}$ and $S'_{sa}$) using an unsupervised clustering technique (Agglo\_Hierarchical).

Now, we classify the 1100 pages into 3 classes ($S'_{fb}$, $S'_{mb}$ and $S'_{sb}$) supervised by the controlled database $D_c$. For this classification, we use GoogLeNet due to its promising performance in other computer vision related tasks \cite{31}. Here also, we arbitrarily choose $n_p$ ($= 400$, fixed empirically) number of text patches in a page and input to the GoogLeNet. 
A page is classified into a certain class where the 
majority %maximum number 
of its patches fall.

To assess the efficiency of the GoogLeNet on intra-variable handwriting, we perform a 3-class classification experiment on $D_c$ to classify in $S_f$, $S_m$, $S_s$ sets/classes due to having the appropriate ground-truth. For this, we divide $D_c$ into training, validation, and test sets in the ratio of 2:1:1. The performance on the test set of $D_c$ for this 3-class classification problem is 96.87\%, which is quite satisfactory for our task.

For classification of uncontrolled handwritten samples (1100 pages), we perform training on the entire controlled database $D_c$ and test on these 1100 pages. After this classification, we have obtained the members of 3-classes $S'_{fb}$, $S'_{mb}$ and $S'_{sb}$. 

Our intention is to generate 3 sets ($S'_{f}$, $S'_m$ and $S'_s$) of samples containing intra-variable handwriting of an individual. Therefore, each of these 3 sets must contain handwriting samples of every writer. 

From the unsupervised clustering, we obtain $S'_{fa}$, $S'_{ma}$ and $S'_{sa}$ sets. From supervised classification, we get the sets $S'_{fb}$, $S'_{mb}$ and $S'_{sb}$. 
From 153 writers, we choose a certain writer who has at least 2 handwriting samples in each of the $S'_{fa}$, $S'_{ma}$, $S'_{sa}$ sets and the same 2 samples in each of the corresponding $S'_{fb}$, $S'_{mb}$, $S'_{sb}$ sets. Finally, these $3 \times 2$ samples of a writer are put in $S'_{f}$, $S'_m$ and $S'_s$ sets, respectively. Out of 153 writers, this constraint is successfully satisfied by 104 writers. Furthermore, we take advice from some handwriting experts and finally choose 100 writers from the 104 writers.

Our uncontrolled database ($D_{uc}$) contains 3 sets ($S'_{f}$, $S'_m$ and $S'_s$) of intra-variable handwriting samples of 100 writers. Each of the $S'_{f}$, $S'_m$ and $S'_s$ sets contain 2 samples per writer. Therefore, similar to the controlled database $D_c$, this uncontrolled database $D_{uc}$ also contains 600 pages in total. The ratio of male to female writers in $D_{uc}$ is 31:19.

In this uncontrolled database, we observe that the handwriting of most students becomes structurally more distorted and unadorned in the latter pages of the answer booklet.

%% file: 4preprocessing.tex
%4. Preprocessing
\section{Preprocessing}
\label{4preprop}

All the handwritten pages were scanned by a flat-bed scanner at 300 ppi (\emph{pixels per inch}) in 256 gray-values to obtain digital document images. In the preprocessing stage, we label the components of a handwritten document image using a relatively faster single-pass connected component labeling algorithm \cite{7}. The text region is extracted after removal of the non-text components if present any, using the method of \cite{8}. In the text region, the struck-out texts are also deleted by employing the method of \cite{40}, since the presence of struck-out text impedes the usual writer identification performance \cite{4}. However, the style of strike-out strokes \cite{40} may be utilized for writer inspection, which is out of the scope of our current work. Very small sized components such as dots, dashes, commas, colons etc., and noise are also removed. The text-lines and words are segmented using an off-the-shelf 2D Gaussian filter-based method \emph{GOLESTAN-a}, as discussed in \cite{10}. Character level segmentation is also performed using a water reservoir principle-based method \cite{11}.

%% file: 5handcraftedFeature.tex
% 5. Handcrafted Feature Extraction
\section{Handcrafted Feature Extraction}
\label{5HCfeature}

A writer identification task can be viewed as a multi-class classification problem, where the task is to assign the writer-id to the unknown handwritten specimens. Similarly, writer verification can be perceived as a binary classification problem where the task is to answer \emph{yes}/\emph{no} to a questioned handwritten sample as to whether it has been written by a particular writer. The features used for these tasks are described in this section and the following section. We employ both handcrafted features and auto-derived features \cite{52}. Handcrafted features are required to be predesigned explicitly in the traditional way, whereas auto-derived features do not have any explicit design.

The extracted handcrafted features are discussed as follows.

%5.1. Macro-Micro Features (FMM):
\subsection{Macro-Micro Features ($F_{MM}$)}

The macro and micro features of Srihari et al. \cite{12} are quite popular 
since those were very effective in 
%for effective 
writer identification from handwritings of 1500 U.S. population having various ethnic groups/ages/genders. 
Here, we adopt this set of features for our task. 

Initially, we choose the macro feature vector, described in \cite{12}, which contains 11 features: gray-level entropy ($f_1$), gray-level threshold ($f_2$), count of black pixels ($f_3$), interior/exterior contour connectivity ($f_4$-$f_5$), vertical/negative/positive/horizontal contour slope ($f_6$-$f_9$), average slant and height of the text-line ($f_{10}$-$f_{11}$). From \cite{12}, we note that the features $f_1$, $f_2$, and $f_3$ are related to the pen pressure. Here, $f_4$ and $f_5$ reveal the writing movement. The features $f_6$, $f_7$, $f_8$, and $f_9$ are related to stroke formation. The feature $f_{10}$ represents the writing slant and $f_{11}$ is related to the text proportion.

Two paragraph-level macro features are also considered: height to width ratio of a paragraph, i.e., aspect ratio ($f_{12}$) and margin width ($f_{13}$). Three more word-level macro features are also employed, which are upper zone ratio ($f_{14}$), lower zone ratio ($f_{15}$) and length ($f_{16}$). We calculate these paragraph and word-level features over a page and take the average value.

The character-level micro features contain 192-bit gradient, 192-bit structural and 128-bit concavity features, concatenated into a 512-bit feature. The detailed description of these features can be found in \cite{12}. We modify this micro feature slightly to get a page-level feature vector. From a page, we obtain the histogram of this 512-bit feature and normalize it by the character count.

The macro and micro features are concatenated to generate the feature vector $F_{MM}$.

% 5.2. Contour Direction and Hinge Features (FDH):
\subsection{Contour Direction and Hinge Features ($F_{DH}$):}

For writer identification, stroke direction and curvature-based features have been reported to work well \cite{1, 53}. Therefore, we use here the famous contour direction and hinge distribution of handwritten strokes proposed by Bulacu and Schomaker \cite{13}.

Along the writing stroke contour, an angle ($\phi$) histogram is generated and normalized into a probability distribution $p_f(\phi)$. From the horizontal direction, the angle ($\phi$) is calculated as: 

\begin{equation}
 \phi=tan^{-1}\frac{y_{i+\epsilon}-y_i}{x_{i+\epsilon}-x_i};
\end{equation}
%(1)
where, $x_i$ and $y_i$ denote the row and column indices of the $i^{th}$ object pixel. The $\epsilon$ depends on stroke-thickness and is fixed as 5 in \cite{13}. 
In our task, $\epsilon ~(>1)$ is data-driven, and worked well 
for 
%when seen empirically. In this case, 
$\epsilon = \max(2,\lfloor{\mu_{sw}-\sigma_{sw}}\rfloor)$, 
where $\mu_{sw}$ is the average stroke-width and $\sigma_{sw}$ is the standard deviation of stroke-width in a page. The number of histogram bins ($n_b$) is set as 12 within the range of $0^o-180^o$. Hence, $15^o$ per bin is engaged. Clearly, the dimension of this feature $p_f(\phi)$ or $f_{cd}$ is 12.

In \cite{13}, for the contour hinge feature $f_{ch}$, two contour fragments, joined to a common end, making angles $\phi_1$ and $\phi_2$ (where, $\phi_2 \geq \phi_1$), spanning all four quadrants ($360^o$), are considered. A normalized histogram is generated with a joint probability distribution $p_f(\phi_1,\phi_2)$.
Similar to \cite{13}, the number of histogram bins ($n_b$) is set to 12, leading to $n_b(2n_b + 1) = 300$-dimensional feature vector. 

By concatenating $f_{cd}$ and $f_{ch}$, we obtain $F_{DH}$.

% 5.3. Direction and Curvature Features at Keypoints (FDC):
\subsection{Direction and Curvature Features at Keypoints ($F_{DC}$)}
\label{sec.5.3}

We intend to ascertain some similarities between handwriting specimens of an individual. Therefore, we focus on some points of interest, i.e., \emph{keypoints} ($\rho_i$), on the handwritten strokes. These keypoints are obtained by combining some structural points (i.e., start/end, branch and curved points) and SIFT (\emph{Scale-Invariant Feature Transform}) keypoints on Bengali handwritten ink-strokes, as described in \cite{6}.

Here, our plan is to observe the movement of writing strokes on these keypoints, and therefore, we capture the stroke direction and curvature at the keypoints. For \emph{direction} and \emph{curvature} feature extraction from \emph{offline} handwritten strokes, we use the idea of ``The NPen++ Recognizer'' \cite{15} which deals with \emph{online} handwriting.

We calculate the writing \emph{direction} between two connected keypoints $\rho_i$ and $\rho_{i+1}$ in terms of Cosine and Sine values and use them as features $f_{dc}$ and $f_{ds}$, respectively.
\begin{equation}
 f_{dc}\equiv cos(\theta_i)=\frac{\rho_{i+1}.x-\rho_i.x}{d_i}
\end{equation}
\begin{equation}
 f_{ds}\equiv sin(\theta_i)=\frac{\rho_{i+1}.y-\rho_i.y}{d_i}
\end{equation}
where, $\rho_i.x$ and $\rho_i.y$ are the row and column indices of $\rho_i$, and \\ 
\noindent
$ d_i = \sqrt{(\rho_{i+1}.x-\rho_i.x)^2 + (\rho_{i+1}.y-\rho_i.y)^2} $ .

The \emph{curvature} of a writing stroke is the angle made by the line fragments 
$\overline{\rho_{i-1}~\rho_i}$ (from $\rho_{i-1}$ to $\rho_i$) 
and $\overline{\rho_i~\rho_{i+1}}$  (from $\rho_i$ to $\rho_{i+1}$). 
The Cosine and Sine values of this angle are calculated and employed as features $f_{cc}$ and $f_{cs}$, respectively.
\begin{equation}
 f_{cc} \equiv cos(\theta_i - \theta_{i-1}) = cos{\theta_i}cos{\theta_{i-1}} + sin{\theta_i}sin{\theta_{i-1}}
\end{equation}
\begin{equation}
 f_{cs} \equiv sin(\theta_i - \theta_{i-1}) = sin{\theta_i}cos{\theta_{i-1}} - cos{\theta_i}sin{\theta_{i-1}}
\end{equation}

For each of these four features ($f_{dc}$, $f_{ds}$, $f_{cc}$, $f_{cs}$), we generate separate normalized histograms spanning the range of [-1, 1] for a number of bins $n_b = 200$. Therefore, the dimension of each feature vector is 200.

%Combining 
Concatenating
features $f_{dc}$, $f_{ds}$, $f_{cc}$ and $f_{cs}$, we get $F_{DC}$.

%% file: 6autoDerivedFeature.tex
%6. Auto-derived Feature Extraction
\section{Auto-derived Feature Extraction}
\label{6ADfeature}

The auto-derived features are mainly extracted using a convolutional neural network (CNN). The convolutional architecture generally contains two parts: front and rear. The front part typically extracts the features. The rear part is used for classification 
(refer to Section \ref{7.2.ADfWI}).

The front part of the CNN takes an input image. We use some patch-based strategies to feed fixed sized input images \cite{18}. Here, we do not use any image normalization, since it impedes the writer identification performance \cite{17}. The patch selection is not performed through the classical sliding-window technique, since the text-lines are not skew-normalized. Sliding a window horizontally through the middle of the text-line (main text-body height) is conceivable, but the information may be lost for several cases such as for highly skewed text-lines, for overlapping text-lines with lesser inter-text-line gaps, etc. 

Here, two types of patches are selected as follows.

(a) \emph{patch\textsubscript{char}}:  %$patch_{char}$
We already have the character-level information from the pre-processing stage. We find the center of gravity (\emph{$p$\textsubscript{\tiny{CG}}}) of a segmented character image. 
Then we take a $n_{char} \times n_{char}$ window centering the \emph{$p$\textsubscript{\tiny{CG}}}, and consider it as a character-level patch, say patch\textsubscript{char}.

(b) \emph{patch\textsubscript{allo}}: 
We have obtained some keypoints on writing strokes, as mentioned in Section \ref{sec.5.3}. 
A neighboring window of size $n_{allo} \times n_{allo}$ centered at a keypoint is used as a patch. This patch is an allographic-level patch, say patch\textsubscript{allo}.

From a text sample, all the patch\textsubscript{char}s and patch\textsubscript{allo}s are extracted. Each patch\textsubscript{char} is fed to the front part of the CNN and a feature vector $f_{pc}$ is obtained. Similarly, for each patch\textsubscript{allo}, a feature vector $f_{pa}$ is generated.

In our task, the following deep-learning architectures are used separately for patch\textsubscript{char} and patch\textsubscript{allo} as inputs.

% 6.1. Basic\_CNN: 
\subsection{Basic\_CNN}

\begin{figure}
  \centering
   \includegraphics[width=0.7\linewidth]{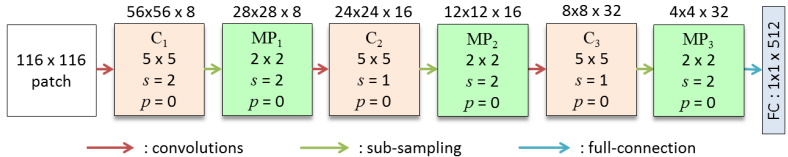}
  \caption{BCNN\textsubscript{char}: Basic\_CNN architecture as a feature extractor while using patch\textsubscript{char}.}
  \label{fig:fig3}
 \end{figure}

The LeNet-5 is a celebrated convolutional network that works well on various machine learning problems \cite{19}. Our Basic\_CNN architecture is primarily influenced by this LeNet-5 and provides an initial flavor of a deep learning model for our task. Here, we use 3 convolutional layers (C\textsubscript{i}), each followed by a sub-sampling (max-pooling, MP\textsubscript{i}) layer. The used feature map count with map size, filter size, stride ($s$), padding ($p$) values are shown in Fig. \ref{fig:fig3}. For example, the first convolutional layer (C\textsubscript{1}) contains 8 feature maps of size $56 \times 56$ each, and each feature map is connected to a $5 \times 5$ neighbor window of the input. Here, $s = 2$ and $p = 0$ is used. 
For each convolutional layer, instead of the \emph{tanh} activation function  of LeNet-5, we use ReLU (\emph{Rectified Linear Unit}) \cite{23} due to its advantages of sparsity and reduced likelihood of vanishing gradient.

For feature extraction using both patch\textsubscript{char} and patch\textsubscript{allo}, the used Basic\_CNN architectures (BCNN\textsubscript{char} and BCNN\textsubscript{allo}) are almost similar except some minor differences. The patch\textsubscript{char} size is $n_{char} \times n_{char}$, and patch\textsubscript{allo} size is $n_{allo} \times n_{allo}$. Here, $n_{allo} = \lfloor{n_{char}/2}\rfloor$  is used. We fix the $n_{char}$ as 116. 
For feeding patch\textsubscript{char} in the first convolutional layer (C\textsubscript{1}) of BCNN\textsubscript{char}, $s=2$ and $p=0$ are used. 
However, for C\textsubscript{1} of BCNN\textsubscript{allo}, $s=1$ and $p=1$ are employed. The rest of the BCNN\textsubscript{allo} architecture is kept similar to BCNN\textsubscript{char} (refer to Fig. \ref{fig:fig3}). In this case, we use \emph{Stochastic Gradient Descent} (SGD) as optimizer with \emph{initial\_learning\_rate} = 0.01, \emph{momentum} = 0.9, and \emph{weight\_decay} = 0.0005. The other parameters of Basic\_CNN are similar to the LeNet-5 \cite{19}.  

Employing this Basic\_CNN, we obtain a feature vector of size 512 for each patch.

% Fig. \ref{fig:fig3} Basic\_CNN architecture as a feature extractor while using patch\textsubscript{char}.

 \begin{figure}[!b]
  \centering
   \includegraphics[width=0.45\linewidth]{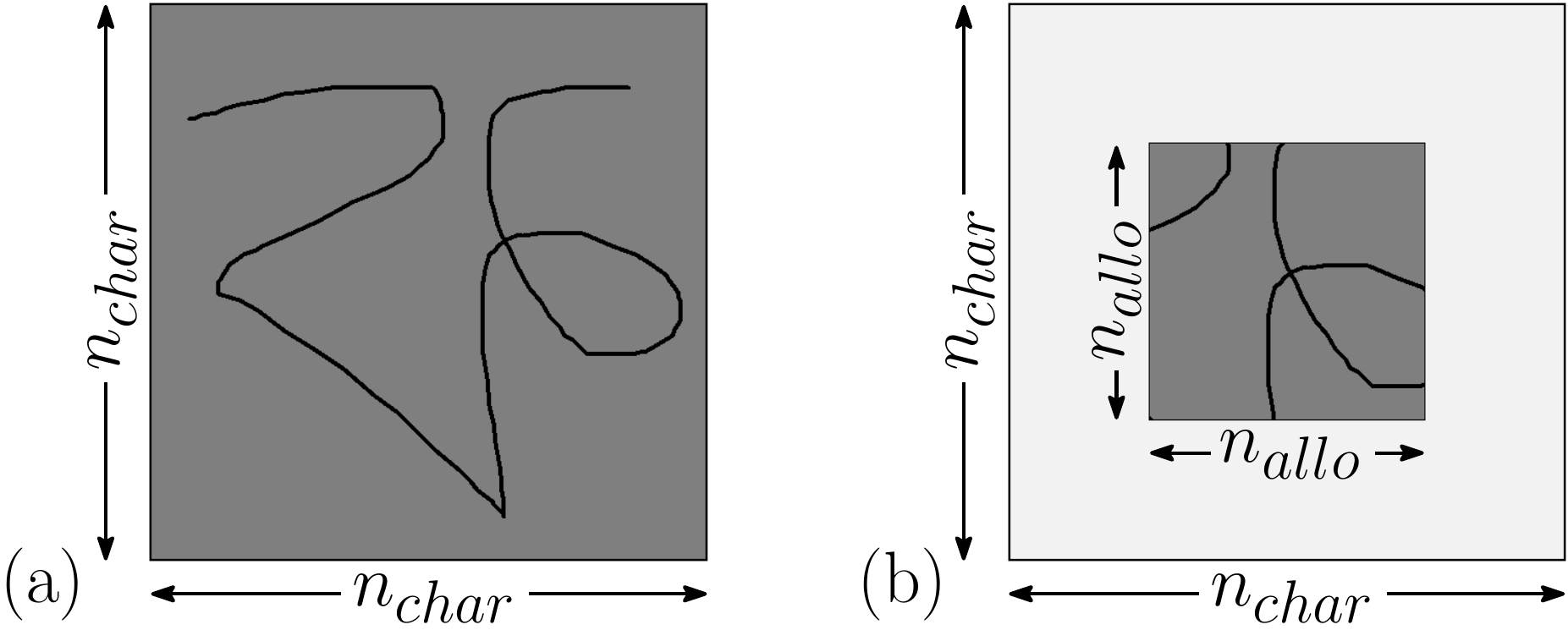}
%    (a)~\includegraphics[width=0.25\linewidth]{fig/xa.eps}
%    ~~~(b)~\includegraphics[width=0.25\linewidth]{fig/xb.eps}
\caption{(a) patch\textsubscript{char} of size $n_{char} \times n_{char}$, 
(b) patch\textsubscript{allo} of size $n_{allo} \times n_{allo}$ is shown in dark-gray, 
and the zero-padding, bounding the patch\textsubscript{allo} is shown in light-gray color.}
  \label{fig:figx}
 \end{figure}

% 6.2. SqueezeNet: 
\subsection{SqueezeNet}
\label{6.2.SqueezeNet}

The AlexNet \cite{43} is one of the pioneering models of deep learning revolution and the winner of ILSVRC (\emph{ImageNet Large Scale Visual Recognition Challenge})-2012 \cite{9}. The recent deep learning era has been started from AlexNet \cite{20}. For our task, we employ some major deep learning architectures, discussed here and the following subsections.

The SqueezeNet architecture provides AlexNet-level accuracy with lesser parameters and a reduced demand on memory \cite{30}. Therefore, we use SqueezeNet here, instead of AlexNet. The details of the SqueezeNet can be found in \cite{30}, and we use the \emph{Simple Bypass} version of this network. Here, we call the layers by their names as used in \cite{30}. For our task, we use the same weights trained on ImageNet \cite{9, 43} by adopting the concept of \emph{transfer learning} \cite{69}.

We select $n_{char} \times n_{char}$ sized patch\textsubscript{char} and $n_{allo} \times n_{allo}$ sized patch\textsubscript{allo} to be fed separately to the SqueezeNets (SN\textsubscript{char} and SN\textsubscript{allo}), where $n_{allo} = \lfloor{n_{char}/2}\rfloor$. The SqueezeNet takes a standard input of fixed size, i.e., 224 $\times$ 224. Here, $n_{char}$ equals to 224 and consequently, $n_{allo}$ becomes 112. Therefore, we use a zero-padding of width 56 ($=\lfloor(n_{char} - n_{allo})/2\rfloor$) to the boundary of patch\textsubscript{allo}, as shown in Fig. \ref{fig:figx}, for maintaining the standard input size of SqueezeNet.

After the \emph{conv10} and \emph{avgpool10} (refer to \cite{30}) layer of SqueezeNet, we obtain a 100 (number of writers)-sized feature vector for each patch.

% 6.3. GoogLeNet:
\subsection{GoogLeNet}

We choose this network, since it won the ILSVRC-2014 \cite{9} competition and obtained a performance closer to human-being. The details of this architecture can be found in \cite{31}. 
This GoogLeNet architecture is also called as \emph{Inception V1}. We employ two separate GoogLeNets (GN\textsubscript{char} and GN\textsubscript{allo}) to feed patch\textsubscript{char} and patch\textsubscript{allo}. The input patch size is similar to the SqueezeNet. The size of patch\textsubscript{char} is $224 \times 224$ and patch\textsubscript{allo} is $112 \times 112$. 
Here also, patch\textsubscript{allo} is bound by a zero-padding of width 56. The rest of the GN\textsubscript{char} and GN\textsubscript{allo} architectures are similar to the GoogLeNet of \cite{31}. 
The weights are transferred by pre-training of GoogLeNet using ImageNet data.

% After the \emph{avg pool} layer \cite{31}, we obtain 1024-sized feature vector.
After the \emph{avg pool} layer \cite{31}, a 1024-dimensional feature vector can be obtained.

% 6.4. Xception Net:
\subsection{Xception Net}

The refined versions of GoogLeNet (\emph{Inception V1}) \cite{31} are \emph{Inception V2} \cite{32} and \emph{Inception V3} \cite{33}. 
Also, the Xception Net \cite{34} is a stronger version of the \emph{Inception V3}. Therefore, we use this Xception Net (with two fully connected layers). The name ``Xception'' is coined from ``Extreme Inception''. 

The patch\textsubscript{char} fed Xception Net (XN\textsubscript{char}) takes $n_{char} \times n_{char}$ sized input images and follows the architecture of \cite{34}. Xception Net \cite{34} takes $299 \times 299$ sized input image. Therefore, we use $n_{char} = 299$. A separate Xception Net (XN\textsubscript{allo}) is used to feed patch\textsubscript{allo} of size $n_{allo} \times n_{allo}$. Here, we use $n_{allo} = \lfloor{n_{char}/2}\rfloor$ = 149. Since, Xception Net takes fixed sized input of 299 $\times$ 299, a zero-padding of width 75 ($=\lfloor(n_{char} - n_{allo})/2\rfloor$) is employed here, similar to the scheme used for SqueezeNet (refer to Section \ref{6.2.SqueezeNet}, Fig. \ref{fig:figx}). 
The pre-trained Xception Net on ImageNet data by the transfer learning \cite{34} is used here with the same weights.

We obtain a 2048-dimensional feature vector from the \emph{GlobalAveragePooling} layer \cite{34} of both XN\textsubscript{char} and XN\textsubscript{allo}.

% 6.5. VGG-16:
\subsection{VGG-16}

The VGG architecture was the runner-up of the competition ILSVRC-2014. We choose the 16 layers' VGG architecture due to its simplicity and uniformity in convolutions. The detail of this architecture is reported in \cite{36}.

We use two VGG-16 networks (VN\textsubscript{char} and VN\textsubscript{allo}). The VGG-16 takes a fixed size input of $224 \times 224$. Therefore, we input $224 \times 224$ sized patch\textsubscript{char} to the VN\textsubscript{char}. 
Similar to the SqueezeNet, here also, we use $112 \times 112$ sized patch\textsubscript{allo} with 56 pixel wide zero-padding to feed to the VN\textsubscript{allo}. Otherwise, VN\textsubscript{allo} and VN\textsubscript{char} networks are the same, and follow the architecture of VGG-16 pre-trained on ImageNet data \cite{36}.

After the \emph{FC-4096} layer \cite{36}, we obtain a 4096-dimensional feature vector from each of VN\textsubscript{allo} and VN\textsubscript{char}.

% 6.6. ResNet-101:
\subsection{ResNet-101}

This architecture won ILSVRC-2015 and beat human-level performance on ImageNet data \cite{9}. 
Although ResNet is very deep, it is faster and has fewer parameters compared to the VGG network. The novelty of the ResNet (\emph{Residual Network}) is its residual or skip connections. The details of this architecture can be found in \cite{35} and we use the ResNet with 101 layers. 
Here, we use two such nets (say, RN\textsubscript{char} and RN\textsubscript{allo}). 

The ResNet also takes fixed sized, i.e., $224 \times 224$ input. Therefore, we feed $224 \times 224$ sized patch\textsubscript{char} as input to the RN\textsubscript{char}. Similar to the SqueezeNet, we feed $112 \times 112$ sized patch\textsubscript{allo} with zero-padding of width of 56 to the RN\textsubscript{allo} (refer to Section \ref{6.2.SqueezeNet}). The rest of the RN\textsubscript{allo} is similar to the RN\textsubscript{char}, and both of them follow the architecture of ResNet-101 as reported in \cite{35}. The weights are transferred by pre-training on ImageNet database \cite{9, 35}.

For each of the RN\textsubscript{char} and RN\textsubscript{allo}, we obtain a 2048-dimensional feature vector after the \emph{avg pool} layer \cite{35}.

%% file: 7writerIdentification.tex
% 7. Writer Identification
\section{Writer Identification}
\label{7WI}

As discussed earlier in Section \ref{intro}, the writer identification problem is a multi-class classification problem, where the number of classes is equal to the total count of writers.

% 7.1. Handcrafted Feature-based Identification:
\subsection{Handcrafted Feature-based Identification}

The handcrafted feature vector obtained from a text sample is fed to an SVM classifier to mark the text sample to its writer-id. The SVM generally works well for multi-class classification in a wide range of pattern recognition applications \cite{37}. 
With regards to the SVM-based multi-class classification for handwriting-related tasks, the \emph{one-against-all} strategy works better than the \emph{one-against-one} \cite{38}. 
It can also be noted from \cite{39, 40} that the SVM with an RBF (\emph{Radial Basis Function}) kernel \cite{41} works better than some other classifiers such as k-NN (k-\emph{Nearest Neighbors}), MLP (\emph{Multi-Layer Perceptron}), MQDF (\emph{Modified Quadratic Discriminant Function}) and SVM-linear for Abjad (Farsi), Alphabetic (English) and Abugida (Bengali) handwritings. Hence, we use the one-against-all SVM-RBF for our task.

The SVM-RBF hyper-parameters $\cal{C}$ and $\gamma$ are essential to be tuned to avoid overfitting and to regulate the decision boundary, respectively \cite{42}. For optimal performance of the classifier, the hyper-parameters are selected from a tuning set. We use the traditional grid-searching technique for this purpose \cite{41}. A suitable value for $\cal{C}$ is chosen at first from a range of values by cross-validation and then several $\gamma$'s are tested from a range of values for better $\cal{C}$'s. 

The best performance is obtained for $\cal{C}$ = ${2}^{2}$ within the range [$2^{-3}, 2^{-2}, \dots, 2^6$] 
and $\gamma = 2^4$ within the range [$2^{-3}, 2^{-2}, \dots, 2^8$]. Here, 5-fold cross-validation is used.

% 7.2. Auto-derived Feature-based Identification:
\subsection{Auto-derived Feature-based Identification}
\label{7.2.ADfWI}

From the text samples, writers are classified using the rear part of the convolutional neural architectures.

For Basic\_CNN, the rear classifier part is actually an MLP with 1 hidden layer containing 256 nodes, set empirically. 
The output layer's nodes depict the number of writer classes. 
%The output layer contains the number of nodes equal to the count of writer classes.
%Here, the training is performed with scaled conjugate gradient backpropagation since it requires less memory. The MSE (\emph{Mean Squared Error}) is employed as the performance parameter. The engaged activation function is the hyperbolic-tangent sigmoid. 

The rear part of SqueezeNet, GoogleNet, Xception Net, VGG-16, ResNet-101 are used for writer identification (classification). The rear parts of the SqueezeNet, GoogleNet, Xception Net, VGG-16, ResNet-101 commence after the \emph{avgpool10} \cite{30}, \emph{avg pool} \cite{31}, \emph{GlobalAveragePooling} \cite{34}, \emph{FC-4096} \cite{36}, \emph{avg pool} \cite{35} layers, respectively. All these respective rear part classifiers follow their original architecture \cite{30, 31, 34, 36, 35}.

We have obtained the features from multiple patches of a handwritten page. 
% In our case, the number of patches in a page is greater than the number of writers of the database. 
Now, to identify a writer on a whole page, the following two strategies are used. The inputs of the classifiers are also based on the following two strategies.

%========================

 \begin{figure}[!b]
  \centering
   \includegraphics[width=\linewidth]{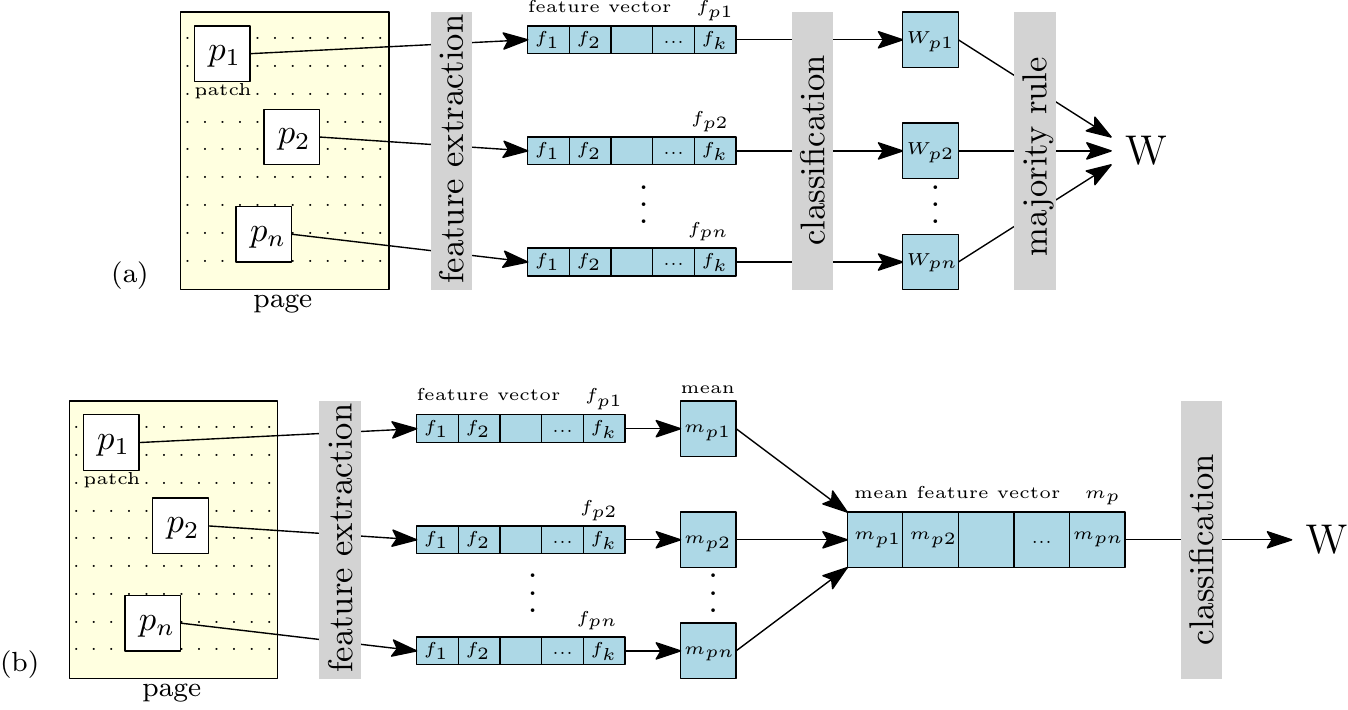} 
%    (b)\includegraphics[width=\linewidth]{fig/mean.pdf}
  \caption{Writer identification strategies: (a) \emph{Strategy-Major}, (b) \emph{Strategy-Mean}.}
  \label{fig:fig4}
 \end{figure}
%Fig. 4. Writer identification strategies: (a) Strategy-Major, (b) Strategy-Mean.

%=======================

(a) \emph{Strategy-Major}: On the basis of feature vector ($f_{pi}$) extracted from each patch ($p_i$), we classify the writer ($W_{pi}$) individually on each patch. In other words, we label each of the multiple patches of a page with a writer-id.

Next, we apply \emph{majority rule} to find the ultimate writer (W) of the page. 
For example, on a page, if the majority of the text patches are marked with writer-A, 
% For example, on a page, if the majority of the text patches are marked as writer-A with more than a certain confidence, 
% For example, on a page, if the highest number of the text patches are marked as writer-A with more than a certain confidence, 
then the overall page is considered as written by writer-A.
This strategy is presented in Algorithm \ref{algo:algo1}.

%====================================
 \begin{algorithm}
   \caption{\emph{Strategy-Major}}
   \begin{algorithmic}[1]
    \State Input: $f_{p1}, f_{p2}, \ldots, f_{pn} |$ feature vectors of patches $p_1, p_2, \dots, p_n$ in a page;
    \State Output: W $|$ writer of the page;
    \For {$i = 1, 2, \ldots, n$} \Comment{$n$ := number of patches}
      \State W{$_{pi}$} = classify($f_{pi}$); \Comment{W{$_{pi}$} := writer of patch $p_i$}
    \EndFor
    \State W= majority (W$_{p1}$, W$_{p2}$, $\dots$, W$_{pn}$);
   \end{algorithmic}
   \label{algo:algo1}
 \end{algorithm}
 %====================================

(b) \emph{Strategy-Mean}: The individual feature vector ($f_{pi}$) obtained from each of the patches ($p_i$) of a page is extracted. Here, we calculate the arithmetic mean ($m_{pi}$) of a feature vector ($f_{pi}$) obtained from each patch ($p_i$). Therefore, for each patch ($p_i$), we have a single scalar mean value ($m_{pi}$). From the mean values of all patches, we generate a mean feature vector ($m_p$) for a page. This mean feature vector is used to classify a page into writer class (W). In Algorithm \ref{algo:algo2}, we present this strategy.

%====================================
 \begin{algorithm}
   \caption{\emph{Strategy-Mean}}
   \begin{algorithmic}[1]
    \State Input: $f_{p1}, f_{p2}, \ldots, f_{pn} |$ feature vectors of patches $p_1, p_2, \dots, p_n$ in a page;
    \State Output: W $|$ writer of the page;
    \For {$i = 1, 2, \ldots, n$} \Comment{$n$ := number of patches}
      \State $m_{pi}$ = arithmeticMean($f_{pi}$); 
    \EndFor
    \State $m_p$ = $\{m_{p1}, m_{p2}, \ldots, m_{pn}\}$;
    \Comment $m_p = $ feature vector of $m_{pi}$'s
    \State W = classify ($m_p$);
   \end{algorithmic}
   \label{algo:algo2}
 \end{algorithm}
 %====================================

For easy and quick understanding, \emph{Strategy-Major} and \emph{Strategy-Mean} are diagrammatically represented in Fig. \ref{fig:fig4}.

Two types of patches patch\textsubscript{char} and patch\textsubscript{allo} (refer to Section \ref{6ADfeature}) are used in both \emph{Strategy-Major} and \emph{Strategy-Mean} for writer identification.

%% file: 8writerVerification.tex
% 8. Writer Verification
\section{Writer Verification}
\label{8WV}

In this section, we discuss the writer verification task using handcrafted features followed by auto-derived features.

% 8.1. Handcrafted Feature-based Verification:
\subsection{Handcrafted Feature-based Verification}
\label{sec.8.1}

In the writer verification task, we check whether two handwriting specimens are written by the same person or not. In fact, the goal is to find some distance measure between the two handwritten samples. If this distance is greater than a \emph{decision threshold} $T$, then we infer that the samples are different, and the same, otherwise ($\leq T$) \cite{44}.

The distance measure is calculated using the handcrafted features generated from the handwritten sample. We used several distance measures such as \emph{Minkowski} up to $order~ 5$ (\emph{Manhattan} when the $order = 1$, \emph{Euclidean} when $order = 2$), \emph{Bhattacharya}, \emph{chi-square} ($\chi^2$) and \emph{Hausdorff} \cite{45}. Here the chi-square ($\chi^2$) distance worked well for our purpose.

The chi-square distance ($\chi^2_{ij}$) between features obtained from two handwritten samples, i.e., \emph{sample-i} and \emph{sample-j}, are calculated as follows:
\begin{equation}
 \chi^2_{ij}=\sum^N_{n=1}\frac{(f_{in}-f_{jn})^2}{(f_{in}+f_{jn})} ~;
\end{equation}
where, $f_{in}$ and $f_{jn}$ and  are the feature vectors obtained from \emph{sample-i} and \emph{sample-j}, respectively. $N$ denotes the dimension of the feature-vector and $n$ represents the index.

In this writer verification task, two types of error are considered: \emph{False Accept} ($FA$) and \emph{False Reject} ($FR$). The nomenclatures of these errors depict their definition. $FA$ is an error when two documents are falsely accepted as having the ``same'' source (written by the same writer), though actually they are ``different''. $FR$ is the error when two documents are falsely rejected as ``different'' (written by different writers), when in fact they are written by the ``same'' person. 

The error rates $FAR$ (\emph{False Acceptance Rate}) and $FRR$ (\emph{False Rejection Rate}) are calculated empirically by integration (up to/from the decision threshold $T$) of the distribution of distances between handwritten samples from different person $P_D(x)$ and the distribution of distances between samples of same person $P_S(x)$, respectively \cite{13}.
\begin{equation}
 FAR = \int_0^T P_D(x)dx ~, ~~~ FRR = \int_T^{\infty} P_S(x)dx ~. 
\end{equation}

From $FAR$ versus $FRR$ plot, we obtained the $EER$ (\emph{Equal Error Rate}) where $FAR = FRR$. 
The writer verification performance in terms of \emph{accuracy} is obtained as $(1-EER) \times 100\%$.

% 8.2. Auto-derived Feature-based Verification:
\subsection{Auto-derived Feature-based Verification}
\label{sec.8.2}

The Siamese Net \cite{16} performs well for weakly supervised similarity metric learning and is successfully applied on various computer vision tasks, e.g., face verification \cite{50}, person re-identification \cite{21}, geo-localization \cite{70}, etc. Therefore, we use this net for writer verification using auto-derived features. Here, our task is treated as a binary classification to classify two handwritten specimens into ``same'' or ``different'' sourced.

\begin{figure}
  \centering
   \includegraphics[width=0.5\linewidth]{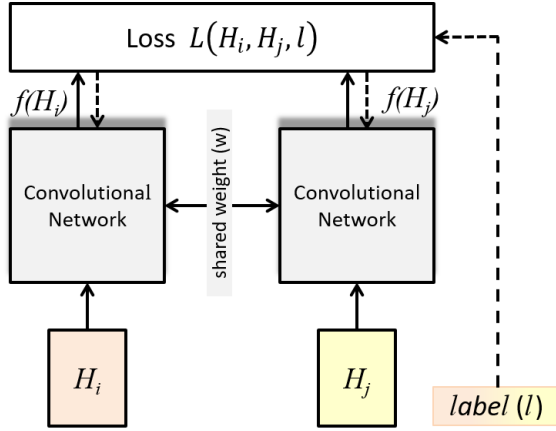}
  \caption{Siamese architecture.}
  \label{fig:fig5}
 \end{figure}

Siamese Net contains identical twin neural architectures to produce two feature vectors from two images to be compared (Fig. \ref{fig:fig5}) \cite{50, 70}. The neural networks of Section \ref{6ADfeature} are used for Siamese twins. Here, employing the \emph{Strategy-Mean} of Section \ref{7.2.ADfWI}, we obtain the mean feature vector from a handwritten page $H_i$, considered as one subnet of the Siamese twins. Parallel to this, another mean feature vector is obtained by the other subnet of the Siamese twins from one more handwritten page $H_j$ to be compared with $H_i$. 
These twin architectures of the Siamese Net are joined by a \emph{loss function} ($L$) at the top to train the similarity metric from the data. In this case, we use a margin-based loss function, i.e., \emph{contrastive loss function} \cite{46}, which is given by:
\begin{equation}
 L(H_i,H_j,l) = \alpha (1-l)D_w^2 + \beta l \{\max(0,m-D_w)\}^2 ~;
\end{equation}
where, \emph{label} $l = 0$, if $H_i$ and $H_j$ are matched as the same, and $l = 1$, otherwise. 
Two constants $\alpha$ and $\beta$ are chosen empirically as $\alpha = 0.5$ and $\beta = 0.5$. 
The margin $m > 0$ is set as the average squared pair distance. 
$D_w \equiv D_w(H_i, H_j) = {\parallel f(H_i) - f(H_j) \parallel}_2$ is the Euclidean distance between $f(H_i)$ and $f(H_j)$ which are two feature vectors generated by mapping of $H_i$ and $H_j$ to a real vector space through the convolutional network.

%Fig. 5. Siamese architecture.

For performance evaluation, a threshold $d$ is used on $D_w$ to verify whether two handwritten samples are written by the ``same'' or a ``different'' writer. All the handwriting pairs $(H_i, H_j)$, inferred to be written by the same writer are denoted as ${\cal{P}}_{same}$, whereas all pairs written by different writers are denoted as ${\cal{P}}_{diff}$. 

Now, we define the set of true positives ($TP$) at $d$ as follows: 
\begin{equation}
 TP(d) = \{ (H_i, H_j) \in {\cal{P}}_{same}, ~with~ D_w(H_i, H_j) \leq d \}~;
\end{equation}
where all the handwriting pairs are correctly classified as the ``same''.

Similarly, when all the handwriting pairs are correctly classified as ``different'', 
then the set of true negatives ($TN$) at $d$ is defined as:
\begin{equation}
 TN(d) = \{ (H_i, H_j) \in {\cal{P}}_{diff}, ~with~ D_w(H_i, H_j) > d \}~.
\end{equation}

The true positive rate ($TPR$) and true negative rate ($TNR$) at $d$ are computed as:
\begin{equation}
 TPR(d) = \frac{|TP(d)|}{|{\cal{P}}_{same}|}~, ~~~ TNR(d) = \frac{|TN(d)|}{|{\cal{P}}_{diff}|}~.
\end{equation}

The overall accuracy is calculated as:
\begin{equation}
 Accuracy = \max\limits_{d \in D_w} \frac{TPR(d)+TNR(d)}{2} ~,
\end{equation}
by varying $d$ in the range of $D_w$ with a step of 0.1.

For writer verification, we use page-level auto-derived features obtained from both patch types 
patch\textsubscript{char} and patch\textsubscript{allo}, using \emph{Strategy-Mean} of Section \ref{7.2.ADfWI}.

%% file: 9experiment.tex
% 9. Experiments and Discussions
\section{Experiments and Discussions}
\label{9Exp}

In this section, at first, we discuss the database employed and the data augmentation for its distribution among the training, validation and test sets. Then we present the results of writer identification and verification.

% 9.1. Database:===========================>
\subsection{Database}

As mentioned in Section \ref{3dataset}, we have generated controlled ($D_c$) and uncontrolled ($D_{uc}$) databases, comprised of 600 pages each. $D_c$ contains 3 sets ($S_f$, $S_m$, and $S_s$) of intra-variable writing. Each of these 3 sets contains 2 handwritten pages of 100 writers. Likewise, $D_{uc}$ contains 3 sets $S'_f$, $S'_m$ and $S'_s$, each having 2 pages by another set of 100 writers.

For intensive experimentation, we need to augment our dataset. The data augmentation technique used here is presented below.

% 9.1.1. Data Augmentation:
\subsubsection{Data Augmentation}

For augmenting our dataset, we are influenced by the idea of ``\emph{DropStroke}'' \cite{47} that is inspired by ``Dropout'' method from the deep neural network \cite{48}. In \cite{47}, the \emph{DropStroke} method is used to generate new data by omitting some strokes randomly from an \emph{online} handwritten Chinese character. 

\begin{figure}
  \centering
   \includegraphics[width=0.2\linewidth]{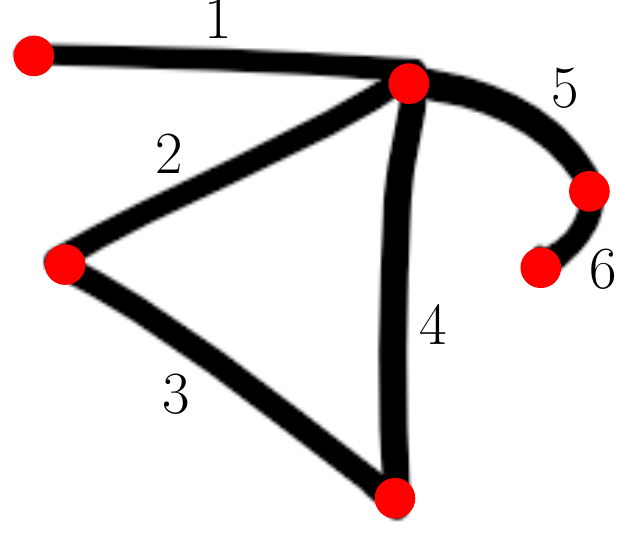}
  \caption{Pictorial representation of a Bengali character component: keypoints are marked by {\textcolor{red}{red}} dots and the six edges are numbered from \textquoteleft{1}' to \textquoteleft{6}'.}
  \label{fig:fig6}
 \end{figure}

Here, the \emph{offline} data lacks the advantage of stroke drawing information of online data. However, we remodel the \emph{DropStroke} as per our requirement for offline handwriting. We use the keypoint information here (refer to Section \ref{sec.5.3}). The ink-pixel connection between two consecutive keypoints is considered as an edge/path/stroke. We drop one edge from a text component (i.e., mostly character, obtained in Section \ref{4preprop}) in such a strategy, so that the number of connected components does not increase. 
Thus, in Fig. \ref{fig:fig6}, edge \textquoteleft{1}' or \textquoteleft{5}' or \textquoteleft{6}' cannot be dropped since it will generate extra component; all the remaining edges can be dropped without any violation of this strategy.

%Fig. 6. Pictorial representation of a Bengali character component: keypoints are marked by red dots and the six edges are numbered from 1 to 6.

To generate new samples from a page, we drop $\lceil{\alpha_d.n_d}\rceil$ number of edges arbitrarily subject to the above condition. Here, $n_d$ is the number of characters in a page and $\alpha_d$ is a parameter in a range of [0.1, 1], set empirically. The value of $n_d$ is computed in Section \ref{4preprop}.  

\begin{figure}[!htb]
  \centering
  \tiny (a){\includegraphics[width=0.47\linewidth]{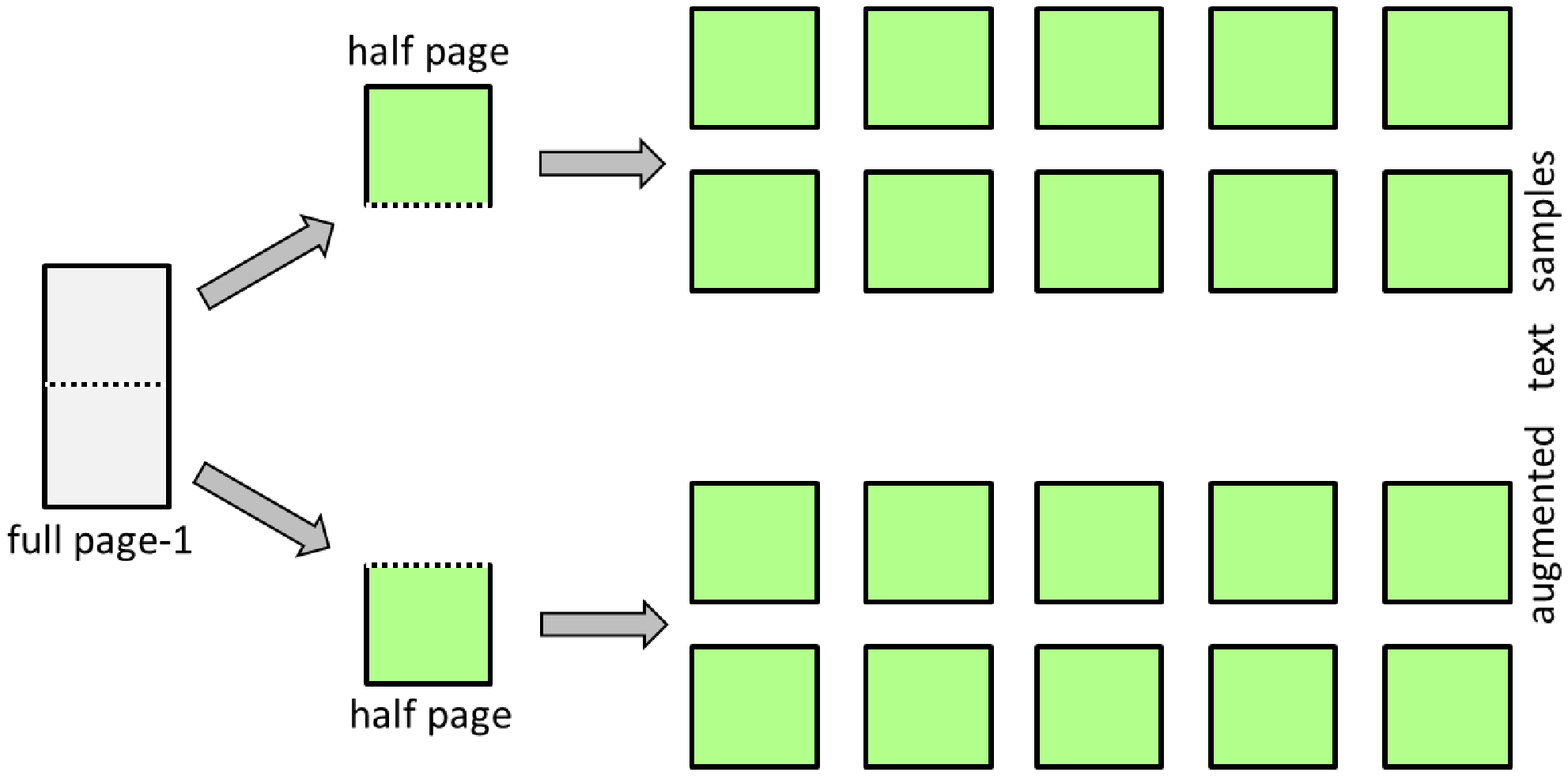}}~%\vline
%    \hrulefill
%    \\~\\
   %-----------------------------------------------------------------------------------------------\\
   (b)\includegraphics[width=0.47\linewidth]{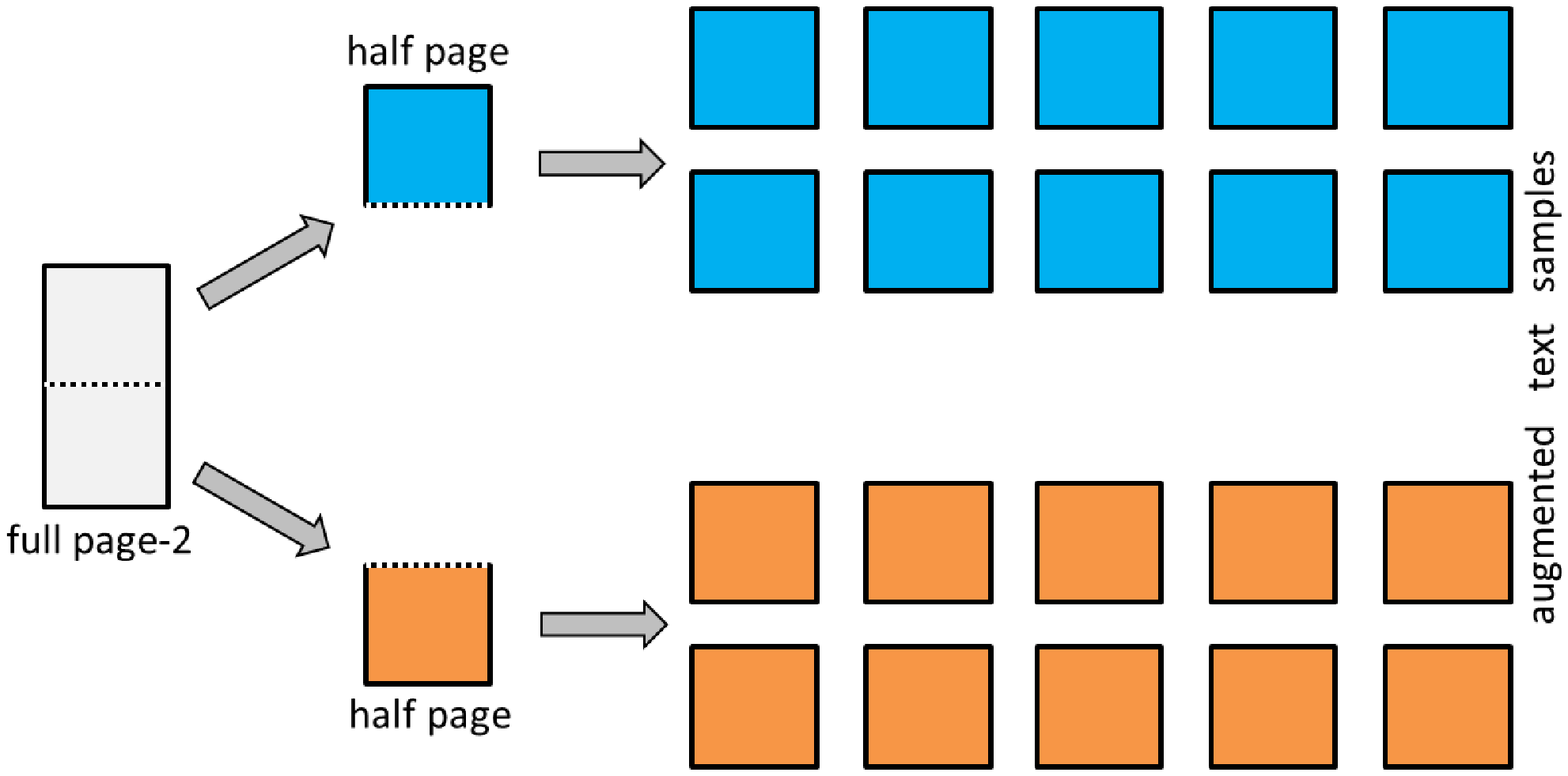}
  \caption{Augmented text samples, generated from 2 handwritten pages of a writer in set $S_f$. The subset $S_{f1}$ contains \textcolor{green1}{green} colored 22 samples for training, $S_{f2}$ contains \textcolor{blue1}{blue} colored 11 samples for validation, and $S_{f3}$ contains \textcolor{orange1}{orange} colored 11 samples for testing.}
  \label{fig:fig7}
 \end{figure}

Initially, a handwritten page is roughly horizontally split into two half-pages, which is a common technique for expansion of data samples in the writer identification task \cite{49}. From each of these two half pages, we generate 10 different samples using our data augmentation technique. Therefore, a handwritten page produces 2 half-pages and 20 ($ = 2 \times 10$) augmented handwritten samples, i.e., overall 22 ($ = 2+20 $) text samples (refer to Fig. \ref{fig:fig7}). 

Thus, each of the databases $D_c$ and $D_{uc}$ contains 13200 (= 600 pages $\times$ 22 samples) text samples. Now, each of the subsets $S_f$, $S_m$, $S_s$ and $S'_f$, $S'_m$, $S'_s$ contains 44 (2 pages $\times$ 22 samples) text samples from each of the 100 writers.

%Fig. 7. Augmented text samples, generated from 2 handwritten pages of a writer in set Sf. The subset $S_{f1}$ contains green colored 22 samples for training, $S_{f2}$ contains blue colored 11 samples for validation, and $S_{f3}$ contains orange colored 11 samples for testing.

We divide both databases $D_c$ and $D_{uc}$ into training, validation, and test sets with a 2:1:1 ratio. 
The set $S_f$ is divided into $S_{f1}$ (training), $S_{f2}$ (validation) and $S_{f3}$ (test) subsets. 
The $S_{f1}$, $S_{f2}$, $S_{f3}$ contains 22, 11, 11 text samples, respectively, from each of the 100 writers. We have ensured distinctiveness among training, validation and test sets, so that no common data is in between any pair of these sets. 
More elaborately, the $S_{f1}$ contains 22 text samples generated from the full \emph{page-1} of a writer in $S_f$. 
The subset $S_{f2}$ contains 11 samples obtained from the top-half of the full \emph{page-2}, 
and $S_{f3}$ contains 11 samples generated from the bottom-half of this \emph{page-2}. 
This is diagrammatically represented in Fig. \ref{fig:fig7} for easy understanding. 
In this case, $S_f = S_{f1} \cup S_{f2} \cup S_{f3}$. 
Similarly, $S_m$, $S_s$ and $S'_f$, $S'_m$, $S'_s$ sets are divided into training, validation and test sets. Here, 
$S_m = S_{m1} \cup S_{m2} \cup S_{m3}$, 
$S_s = S_{s1} \cup S_{s2} \cup S_{s3}$; 
$S'_f = S'_{f1} \cup S'_{f2} \cup S'_{f3}$, 
$S'_m = S'_{m1} \cup S'_{m2} \cup S'_{m3}$, 
$S'_s = S'_{s1} \cup S'_{s2} \cup S'_{s3}$.

%---------------------------------------------------------------------------

Now, for our writer identification/verification task, we train and test with various types of intra-variable data. The experiments are performed in this way to imitate real-life situations, where a particular type of handwriting may be absent (refer to Section \ref{intro}).

% 9.2. Writer Identification Performance:===========================>
\subsection{Writer Identification Performance}
\label{sec.9.2}

In this subsection, we discuss the performance of the writer identification models based on handcrafted features and auto-derived features.

The writer identification accuracy is computed using a ``{Top-$N$}'' criterion, where the correct writer is marked at least one time within the \textquoteleft{$N$}' ($\ll$ total number of writers) top-most classifier output confidences. Here, we compute results of Top-1, Top-2, and Top-5 criteria. However, Top-1 accuracies are presented in more detail for comparison among models. Top-2 and Top-5 accuracies are also presented when a better outcome is achieved.

% 9.2.1. Writer Identification Performance with Handcrafted Features: 
\subsubsection{Writer Identification Performance with Handcrafted Features}
\label{9.2.1}

Here, we discuss the writer identification performance on both databases, $D_c$ and $D_{uc}$, by employing handcrafted features.

% 9.2.1.1. Writer Identification by Handcrafted Features on $D_c$:
\paragraph{\ref{9.2.1}.1. Writer Identification by Handcrafted Features on $D_c$}~\\

The Top-1 writer identification performance on database $D_c$ by employing feature $F_{MM}$ with SVM classifier (say, system ``$WI\_D_c\_F_{MM}$'') is shown in Table \ref{tab:table1}.

\iffalse
%===================================================================================

\begin{table}[!ht]
\scriptsize
\caption{Top-1 writer identification performance of system $WI\_D_c\_F_{MM}$}
\centering
\begin{tabular}{l|c|c|c}
\hline
\multicolumn{1}{c|}{\bf{Set}} & \multicolumn{3}{c}{\bf{Accuracy (\%)}}\\ \hline 
\backslashbox{\bf{Training}}{\bf{Test}} & {\bf{$S_{s3}$}} & {\bf{$S_{m3}$}} & {\bf{$S_{f3}$}}\\ \hline 
{\bf{$S_{s1}$}} & \cellcolor{green!25}61.23 & \cellcolor{red!25}37.17 & \cellcolor{blue!25}29.49\\ \hline 
{\bf{$S_{m1}$}} & \cellcolor{red!25}38.45 & \cellcolor{green!25}60.48 & \cellcolor{yellow!25}36.67\\ \hline 
{\bf{$S_{f1}$}} & \cellcolor{blue!25}31.45 & \cellcolor{yellow!25}34.59 & \cellcolor{green!25}59.32\\ \hline 
{\bf{$S_{s1}+S_{m1}$}} & 61.42 & 61.89 & 37.04\\ \hline 
{\bf{$S_{s1}+S_{f1}$}} & 62.24 & 38.47 & 60.33\\ \hline 
{\bf{$S_{m1}+S_{f1}$}} & 40.72 & 60.95 & 59.86\\ \hline 
{\bf{$S_{s1}+S_{m1}+S_{f1}$}} & \cellcolor{gray!25}62.78 & \cellcolor{gray!25}62.23 & \cellcolor{gray!25}61.31\\ \hline 
\end{tabular}\label{tab:table1}
\end{table}

%===================================================================================
\fi
\iffalse
Table 1. Top-1 writer identification performance of system WI_$D_c$_FMM

\fi

Here, by training with all the training data ($S_{s1}+S_{m1}+S_{f1}$) of $D_c$, and testing only on $S_{s3}$, we obtain a 62.78\% accuracy. Likewise, training on $S_{s1}+S_{m1}+S_{f1}$, and testing on $S_{m3}$ and $S_{f3}$, we obtain 62.23\% and 61.31\% accuracies, respectively.

Experimenting on the same type set yields better accuracy, e.g., training on subset $S_{s1}$ and testing on subset $S_{s3}$ (say, experimental setup $S_{s1}/S_{s3}$ or, $E_{ss}$) provides 61.23\% accuracy, where both training and testing are parts of the set $S_s$. Similarly, experimental setup $S_{m1}/S_{m3}$ ($E_{mm}$) and $S_{f1}/S_{f3}$ ($E_{ff}$) provides a 60.48\% and 59.32\% accuracy, respectively.

Next, we look at the performance of experiments on different training and test sets. For example, training on $S_{s1}$ and testing on $S_{m3}$ (say, experimental setup $S_{s1}/S_{m3}$ or, $E_{sm}$) yields 37.17\% accuracy, which is quite low. 
The reverse experimental setup, i.e., $S_{m1}/S_{s3}$ ($E_{ms}$) also shows poor performance (38.45\% accuracy). Similarly, experimental setups $E_{mf}$, $E_{fm}$, $E_{sf}$, $E_{fs}$ provide poor results, with accuracies 36.67\%, 34.59\%, 29.49\%, 31.45\%, respectively. 
The reason behind such a low outcome is the presence of high variability between the training and test sets. Experimental setups $E_{sf}$ and $E_{fs}$ show the lowest performance, since they contain highly intra-variable writing.

For our task, we define the performance of our model by a tuple of 9 major accuracies (\%) obtained by various experimental setups. This 9-tuple is ($AE_{ss}$, $AE_{mm}$, $AE_{ff}$, $AE_{smv}$, $AE_{sfv}$, $AE_{mfv}$, $AE_{smf/s}$, $AE_{smf/m}$, $AE_{smf/f}$) which is used to compare multiple models used in this paper.  
$AE_{ss}$, $AE_{mm}$, $AE_{ff}$ are the accuracy measures obtained from the $E_{ss}$, $E_{mm}$, $E_{ff}$ experimental setups, respectively. These $AE_{ss}$, $AE_{mm}$, $AE_{ff}$ accuracies show the efficacy of a model on low intra-variable handwriting, which are mostly similar types. In Table \ref{tab:table1}, we show these accuracies highlighted in \colorbox{green!25}{green}, which can be seen better in the softcopy of this paper.

\begin{table}[!t]
\tiny %\scriptsize
\parbox{.45\linewidth}{
\centering
\caption{Top-1 writer identification performance of system $WI\_D_c\_F_{MM}$}
\begin{tabular}{l|c|c|c}
\hline
\multicolumn{1}{c|}{\bf{Set}} & \multicolumn{3}{c}{\bf{Accuracy (\%)}}\\ \hline 
\backslashbox{\bf{Training}}{\bf{Test}} & {\bf{$S_{s3}$}} & {\bf{$S_{m3}$}} & {\bf{$S_{f3}$}}\\ \hline  %\hline
{\bf{$S_{s1}$}} & \cellcolor{green!25}61.23 & \cellcolor{red!25}37.17 & \cellcolor{blue!25}29.49\\ \hline 
{\bf{$S_{m1}$}} & \cellcolor{red!25}38.45 & \cellcolor{green!25}60.48 & \cellcolor{yellow!25}36.67\\ \hline 
{\bf{$S_{f1}$}} & \cellcolor{blue!25}31.45 & \cellcolor{yellow!25}34.59 & \cellcolor{green!25}59.32\\ \hline 
{\bf{$S_{s1}+S_{m1}$}} & 61.42 & 61.89 & 37.04\\ \hline 
{\bf{$S_{s1}+S_{f1}$}} & 62.24 & 38.47 & 60.33\\ \hline 
{\bf{$S_{m1}+S_{f1}$}} & 40.72 & 60.95 & 59.86\\ \hline 
{\bf{$S_{s1}+S_{m1}+S_{f1}$}} & \cellcolor{gray!25}\textbf{62.78} & \cellcolor{gray!25}62.23 & \cellcolor{gray!25}61.31\\ \hline 
\end{tabular}\label{tab:table1}
% \caption{Top-1 writer identification performance of system $WI\_D_c\_F_{MM}$}
}
\hfill
\parbox{.45\linewidth}{
\centering
\caption{Top-1 writer identification performance of system $WI\_D_c\_F_{DH}$}
\begin{tabular}{l|c|c|c}
\hline
\multicolumn{1}{c|}{\bf Set} & \multicolumn{3}{c}{\bf Accuracy (\%)}\\ \hline 
\backslashbox{\bf{Training}}{\bf Test} & $S_{s3}$ & $S_{m3}$ & $S_{f3}$\\ \hline  %\hline
$S_{s1}$ & \cellcolor{green!25}72.67 & \cellcolor{red!25}44.71 & \cellcolor{blue!25}37.29\\ \hline 
$S_{m1}$ & \cellcolor{red!25}45.53 & \cellcolor{green!25}71.86 & \cellcolor{yellow!25}44.17\\ \hline 
$S_{f1}$ & \cellcolor{blue!25}39.23 & \cellcolor{yellow!25}43.59 & \cellcolor{green!25}70.93\\ \hline 
$S_{s1}+S_{m1}$ & 72.92 & 72.84 & 44.85\\ \hline 
$S_{s1}+S_{f1}$ & 72.86 & 45.42 & 71.54\\ \hline 
$S_{m1}+S_{f1}$ & 46.28 & 72.37 & 71.72\\ \hline 
$S_{s1}+S_{m1}+S_{f1}$ & \cellcolor{gray!25}73.49 & \cellcolor{gray!25}73.36 & \cellcolor{gray!25}72.26\\ \hline 
\end{tabular}\label{tab:table2}
% \caption{Top-1 writer identification performance of system $WI\_D_c\_F_{DH}$}
}
\end{table}

$AE_{smv}$ is the average (arithmetic mean) accuracy obtained from experimental setups $E_{sm}$ ($S_{s1}/S_{m3}$) and $E_{ms}$ ($S_{m1}/S_{s3}$). This is depicted with \colorbox{red!25}{red} shade in Table \ref{tab:table1}. 
Similarly, $AE_{sfv}$ is obtained from $E_{sf}$, $E_{fs}$ (\colorbox{blue!25}{blue} shaded in Table \ref{tab:table1}) and $AE_{mfv}$ is obtained from $E_{mf}$, $E_{fm}$ (\colorbox{yellow!25}{yellow} in Table \ref{tab:table1}), respectively. 
$AE_{smv}$, $AE_{sfv}$, $AE_{mfv}$ demonstrate the system performance when both training and test sets contain highly intra-variable writing.

$AE_{smf/s}$ is the accuracy obtained from the experimental setup $E_{smf/s}$, where training is performed on $S_{s1}+S_{m1}+S_{f1}$, and testing is executed on $S_{s3}$. 
$AE_{smf/m}$ and $AE_{smf/f}$ are obtained by testing on $S_{m3}$ and $S_{f3}$, respectively, while training is performed on $S_{s1}+S_{m1}+S_{f1}$, similar to the training of $AE_{smf/s}$. 
Here, $AE_{smf/s}$, $AE_{smf/m}$, $AE_{smf/f}$ show the performance when the system is trained with all available handwriting varieties of an individual. 
In Table \ref{tab:table1}, we show these accuracies in \colorbox{gray!25}{gray} shade.

The performance of the system $WI\_D_c\_F_{MM}$ in terms of 9-tuple is (61.23, 60.48, 59.32, 37.81, 30.47, 35.63, 62.78, 62.23, 61.31).

The Top-1 writer identification performance on database $D_c$ using feature $F_{DH}$ with SVM (say, system ``$WI\_D_c\_F_{DH}$'') is shown in Table \ref{tab:table2}.
The performance of system $WI\_D_c\_F_{DH}$ in terms of 9-tuple is (72.67, 71.86, 70.93, 45.12, 38.26, 43.88, 73.49, 73.36, 72.26).  This can be tallied with Table \ref{tab:table2}, as we tallied $WI\_D_c\_F_{MM}$ performance with Table \ref{tab:table1}.

\iffalse
%===================================================================================

\begin{table}[!ht]
\scriptsize
\caption{Top-1 writer identification performance of system $WI\_D_c\_F_{DH}$}
\centering
\begin{tabular}{l|c|c|c}
\hline
\multicolumn{1}{c|}{\bf Set} & \multicolumn{3}{c}{\bf Accuracy (\%)}\\ \hline 
\backslashbox{\bf{Training}}{\bf Test} & $S_{s3}$ & $S_{m3}$ & $S_{f3}$\\ \hline 
$S_{s1}$ & \cellcolor{green!25}72.67 & \cellcolor{red!25}44.71 & \cellcolor{blue!25}37.29\\ \hline 
$S_{m1}$ & \cellcolor{red!25}45.53 & \cellcolor{green!25}71.86 & \cellcolor{yellow!25}44.17\\ \hline 
$S_{f1}$ & \cellcolor{blue!25}39.23 & \cellcolor{yellow!25}43.59 & \cellcolor{green!25}70.93\\ \hline 
$S_{s1}+S_{m1}$ & 72.92 & 72.84 & 44.85\\ \hline 
$S_{s1}+S_{f1}$ & 72.86 & 45.42 & 71.54\\ \hline 
$S_{m1}+S_{f1}$ & 46.28 & 72.37 & 71.72\\ \hline 
$S_{s1}+S_{m1}+S_{f1}$ & \cellcolor{gray!25}73.49 & \cellcolor{gray!25}73.36 & \cellcolor{gray!25}72.26\\ \hline 
\end{tabular}\label{tab:table2}
\end{table}

%===================================================================================
\fi
\iffalse
Table 2. Top-1 writer identification performance of system WI_$D_c$_FDH.

\fi

The Top-1 writer identification performance on database $D_c$ using feature $F_{DC}$ with SVM (say, system ``$WI\_D_c\_F_{DC}$'') is shown in Table \ref{tab:table3}.
The performance of system $WI\_D_c\_F_{DC}$ in terms of 9-tuple is (71.54, 70.25, 69.71, 43.56, 36.38, 40.80, 72.48, 71.85, 70.93). This can be tallied with Table \ref{tab:table3}.

%===================================================================================

\begin{table}[!ht]
\tiny %\scriptsize
\caption{Top-1 writer identification performance of system $WI\_D_c\_F_{DC}$}
\centering
\begin{tabular}{l|c|c|c}
\hline
\multicolumn{1}{c|}{\bf Set} & \multicolumn{3}{c}{\bf Accuracy (\%)}\\ \hline 
\backslashbox{\bf{Training}}{\bf Test} & $S_{s3}$ & $S_{m3}$ & $S_{f3}$\\ \hline  %\hline
$S_{s1}$ & \cellcolor{green!25}71.54 & \cellcolor{red!25}43.20 & \cellcolor{blue!25}37.04\\ \hline 
$S_{m1}$ & \cellcolor{red!25}43.92 & \cellcolor{green!25}70.25 & \cellcolor{yellow!25}41.06\\ \hline 
$S_{f1}$ & \cellcolor{blue!25}35.72 & \cellcolor{yellow!25}40.54 & \cellcolor{green!25}69.71\\ \hline 
$S_{s1}+S_{m1}$ & 71.75 & 70.84 & 41.74\\ \hline 
$S_{s1}+S_{f1}$ & 71.66 & 43.37 & 70.14\\ \hline 
$S_{m1}+S_{f1}$ & 44.47 & 71.45 & 70.68\\ \hline 
$S_{s1}+S_{m1}+S_{f1}$ & \cellcolor{gray!25}72.48 & \cellcolor{gray!25}71.85 & \cellcolor{gray!25}70.93\\ \hline 
\end{tabular}\label{tab:table3}
\end{table}

%===================================================================================

%===================================================================================

\begin{table}[!ht]
\tiny
\caption{Top-1 writer identification performance using handcrafted features on $D_c$}
\centering
\begin{tabular}{c|c|c|c|c|c|c|c|c|c|c}
\hline
\multirow{2}{*}{\textbf{Model}} & \multicolumn{9}{c|}{\textbf{9-tuple Accuracy (\%)}} & \multirow{2}{*}{\textbf{Rank}} \\ \cline{2-10} 
 & $AE_{ss}$ &  $AE_{mm}$ & $AE_{ff}$ & $AE_{smv}$ & $AE_{sfv}$ & $AE_{mfv}$ & $AE_{smf/s}$ & $AE_{smf/m}$ & $AE_{smf/f}$ & \\ \hline %\hline
$WI\_D_{c}\_F_{MM}$ & 61.23 & 60.48 & 59.32 & 37.81 & 30.47 & 35.63 & 62.78 & 62.23 & 61.31 & 3\\ \hline
$WI\_D_{c}\_F_{DH}$ & 72.67 & 71.86 & 70.93 & 45.12 & 38.26 & 43.88 & 73.49 & 73.36 & 72.26 & 1\\ \hline
$WI\_D_{c}\_F_{DC}$ & 71.54 & 70.25 & 69.71 & 43.56 & 36.38 & 40.80 & 72.48 & 71.85 & 70.93 & 2\\ \hline 
\end{tabular}\label{tab:table4}
\end{table}

%===================================================================================

\iffalse
Table 3. Top-1 writer identification performance of system WI_$D_c$_FDC

\fi

Combining Tables \ref{tab:table1}, \ref{tab:table2} and \ref{tab:table3}, we generate Table \ref{tab:table4} to present only the 9-tuple accuracies of all handcrafted feature-based writer identification models dealing with $D_c$, for comparison and easy visualization. 

The models are ranked using the \emph{Borda count} \cite{51}. Here, the models are initially ranked with respect to each accuracy of the 9-tuple (i.e., performance on each experimental setup), then the aggregate ranking is computed by the \emph{max} rule. If there is a draw between two models, then we provide weightage on the accuracies $AE_{smv}$, $AE_{sfv}$, $AE_{mfv}$. The aggregate ranks are shown in the last columns of 
Tables \ref{tab:table4} - \ref{tab:table7}. 
Here, Rank \textquoteleft{1}' denotes first, i.e., the best performing model, Rank \textquoteleft{2}' indicates the second best performing model, and so on.

\iffalse
x

.
Table 4. Top-1 writer identification performance using handcrafted features on $D_c$.

\fi

Although, we have computed 21 accuracy measures as in Tables \ref{tab:table1} - \ref{tab:table3}, we present the 9-tuple accuracy measures like Table \ref{tab:table4} further for model assessment and for simple visualization.

% 9.2.1.2. Writer Identification by Handcrafted Features on $D_{uc}$:

\paragraph{\ref{9.2.1}.2. Writer Identification by Handcrafted Features on $D_{uc}$}~\\

By employing database $D_{uc}$, here also, we generate three similar handcrafted feature-based writer identification models as mentioned in Section \ref{9.2.1}.1. In Table \ref{tab:table5}, we present the performance of these models on $D_{uc}$ with respect to the 9-tuple accuracy.

In this case, while employing $D_{uc}$, the $AE_{ss}$ accuracy is obtained from an experimental setup $E_{ss}$ where $S'_{s1}$ is used for training and $S'_{s3}$ is used for testing. Similarly, for obtaining $AE_{smf/m}$, training is performed on ($S'_{s1}+S'_{m1}+S'_{f1}$), and testing is executed on $S'_{m3}$. Likewise, other accuracies of 9-tuple are obtained (refer to Section \ref{9.2.1}.1).

%===================================================================================

\begin{table}[!ht]
\tiny
\caption{Top-1 writer identification performance using handcrafted features on $D_{uc}$}
\centering
\begin{tabular}{c|c|c|c|c|c|c|c|c|c|c}
\hline
\multirow{2}{*}{\textbf{Model}} & \multicolumn{9}{c|}{\textbf{9-tuple Accuracy (\%)}} & \multirow{2}{*}{\textbf{Rank}} \\ \cline{2-10} 
 & $AE_{ss}$ &  $AE_{mm}$ & $AE_{ff}$ & $AE_{smv}$ & $AE_{sfv}$ & $AE_{mfv}$ & $AE_{smf/s}$ & $AE_{smf/m}$ & $AE_{smf/f}$ & \\ \hline %\hline
$WI\_D_{uc}\_F_{MM}$ & 60.08 & 58.51 & 57.72 & 37.01 & 29.77 & 33.92 & 61.63 & 61.04 & 60.01 & 3\\ \hline
$WI\_D_{uc}\_F_{DH}$ & 71.79 & 71.21 & 70.19 & 43.85 & 37.01 & 42.21 & 72.93 & 72.66 & 71.97 & 1\\ \hline
$WI\_D_{uc}\_F_{DC}$ & 69.77 & 68.69 & 68.54 & 42.43 & 34.96 & 39.16 & 70.82 & 70.02 & 69.66 & 2\\ \hline 
\end{tabular}\label{tab:table5}
\end{table}

%===================================================================================

\iffalse
x
.
Table 5. Top-1 writer identification performance using handcrafted features on $D_{uc}$.

Model
9-tuple Accuracy (%)
Rank

.
X
\fi

Experimenting on both the databases $D_c$ and $D_{uc}$ using handcrafted features, overall the $F_{DH}$ feature-based model performed the best and the $F_{MM}$ feature-based model achieved the lowest results.

It can be observed from Tables \ref{tab:table4} and \ref{tab:table5} that the accuracies $AE_{smv}$, $AE_{sfv}$, $AE_{mfv}$ are very low. We have noted on $AE_{smv}$, $AE_{sfv}$, $AE_{mfv}$, even Top-2 (Top-5) writer identification performance provides additional at most 0.38\% (2.97\%) and 0.30\% (2.53\%) accuracy on $D_c$ and $D_{uc}$, respectively.

% 9.2.2. Writer Identification Performance with Auto-derived Features:
\subsubsection{Writer Identification Performance with Auto-derived Features}
\label{9.2.2}

We perform writer identification by feeding patch\textsubscript{char} to the Basic\_CNN with the \emph{Strategy-Major}, and call this model: ``BCNN\_char\_major''. 
Likewise, feeding patch\textsubscript{char} to the Basic\_CNN with the \emph{Strategy-Mean}, is called a ``BCNN\_char\_mean'' model. The patch\textsubscript{allo}s when input into the Basic\_CNN with the \emph{Strategy-Major} and the \emph{Strategy-Mean}, are called the ``BCNN\_allo\_major'' and the ``BCNN\_allo\_mean'', respectively.

Thus, a convolutional network produces 4 variations of auto-derived feature-based models for writer identification, i.e., ``x\_char\_major'', ``x\_char\_mean'', ``x\_allo\_major'', and ``x\_allo\_mean''. 
Here, \textquoteleft{x}' is to be replaced by the convolutional network name. 
The \textquoteleft{x}' is replaced by \textquoteleft{SN}', \textquoteleft{GN}', \textquoteleft{XN}', \textquoteleft{VN}', \textquoteleft{RN}' while employing SqueezeNet, GoogLeNet, Xception Net, VGG-16, ResNet-101, respectively. For example, feeding patch\textsubscript{char} in GoogLeNet with the \emph{Strategy-Mean} is called: ``GN\_char\_mean''.

Consequently, 6 types of convolutional networks, each of 4 various configurations, produce a total of 24 
($= 6 \times 4$) models. 
Here, we present the previously mentioned 9-tuple accuracy measure for each model (refer to Section \ref{9.2.1}.1).

% 9.2.2.1. Writer Identification by Auto-derived Features on $D_c$:
\paragraph{\ref{9.2.2}.1. Writer Identification by Auto-derived Features on $D_c$}~\\

In Table \ref{tab:table6}, we present the 9-tuple writer identification accuracies of auto-derived feature-based models performing on the $D_c$ database. 

%===================================================================================

\begin{table}[!b]
\tiny
\caption{Top-1 writer identification performance using auto-derived features on $D_c$}
\centering
\begin{tabular}{rl|l|c|c|c|c|c|c|c|c|c|c}
\cline{3-13}
\multicolumn{1}{c}{}&\multicolumn{1}{c}{}&\multicolumn{1}{c|}{\textbf{Model}} & \multicolumn{9}{c|}{\textbf{9-tuple Accuracy (\%)}} & \multirow{2}{*}{\textbf{Rank}} \\ \cline{4-12} 
 \multicolumn{1}{c}{}&\multicolumn{1}{c}{}&\multicolumn{1}{c|}{($WI\_D_c$)}& $AE_{ss}$ &  $AE_{mm}$ & $AE_{ff}$ & $AE_{smv}$ & $AE_{sfv}$ & $AE_{mfv}$ & $AE_{smf/s}$ & $AE_{smf/m}$ & $AE_{smf/f}$ & \\ \hline %\hline
\parbox[t]{0.01mm}{\multirow{4}{*}{\rotatebox[origin=c]{90}{\tiny{Basic}}}}&
 \parbox[t]{.01mm}{\multirow{4}{*}{\rotatebox[origin=c]{90}{\tiny{\_CNN}}}}& BCNN\_char\_major & 81.25 & 81.72 & 81.51 & 57.58 & 50.03 & 53.67 & 83.72 & 83.32 & 82.58 & 23\\ \cline{3-13}
&&BCNN\_allo\_major & 81.78 & 81.48 & 81.09 & 58.25 & 48.92 & 53.35 & 84.16 & 82.35 & 81.89 & 24\\ \cline{3-13} 
&&BCNN\_char\_mean & 83.56 & 83.12 & 82.37 & 60.02 & 51.27 & 55.93 & 85.58 & 84.73 & 83.89 & 22\\ \cline{3-13} 
&&BCNN\_allo\_mean & 84.53 & 83.63 & 82.92 & 60.59 & 51.34 & 55.25 & 86.21 & 85.33 & 84.74 & 21\\ \hline 
\parbox[t]{.01mm}{\multirow{4}{*}{\rotatebox[origin=c]{90}{\tiny{Squeeze}}}}&
\parbox[t]{.01mm}{\multirow{4}{*}{\rotatebox[origin=c]{90}{\tiny{Net}}}}&SN\_char\_major & 87.27 & 85.54 & 85.14 & 62.92 & 53.15 & 58.03 & 88.16 & 87.69 & 86.29 & 19\\ \cline{3-13} 
&&SN\_allo\_major & 86.75 & 84.55 & 85.37 & 62.25 & 52.71 & 57.35 & 87.25 & 85.56 & 87.02 & 20\\ \cline{3-13} 
&&SN\_char\_mean & 88.46 & 87.79 & 87.31 & 65.35 & 56.79 & 61.53 & 90.27 & 89.35 & 89.05 & 18\\ \cline{3-13} 
&&SN\_allo\_mean & 89.23 & 88.48 & 87.72 & 66.56 & 56.05 & 61.45 & 90.53 & 89.72 & 89.25 & 17\\ \hline 
\parbox[t]{.01mm}{\multirow{4}{*}{\rotatebox[origin=c]{90}{\tiny{GoogLe}}}}&
\parbox[t]{.01mm}{\multirow{4}{*}{\rotatebox[origin=c]{90}{\tiny{Net}}}}&GN\_char\_major & 91.09 & 90.07 & 87.97 & 67.37 & 57.08 & 62.58 & 91.60 & 91.44 & 90.44 & 13\\ \cline{3-13} 
&&GN\_allo\_major & 90.29 & 90.16 & 87.48 & 66.28 & 56.68 & 61.42 & 90.81 & 90.79 & 90.07 & 16\\ \cline{3-13} 
&&GN\_char\_mean & 91.34 & 89.78 & 89.29 & 68.04 & 58.28 & 63.99 & 92.58 & 90.62 & 91.20 & 12\\ \cline{3-13}
&&GN\_allo\_mean & 91.39 & 90.48 & 90.04 & 69.12 & 58.92 & 63.24 & 93.13 & 92.16 & 91.25 & 11\\ \hline 
\parbox[t]{.01mm}{\multirow{4}{*}{\rotatebox[origin=c]{90}{\tiny{VGG}}}}&
\parbox[t]{.01mm}{\multirow{4}{*}{\rotatebox[origin=c]{90}{\tiny{-16}}}}&VN\_char\_major & 91.02 & 89.25 & 88.58 & 67.33 & 57.53 & 62.22 & 91.71 & 90.03 & 90.43 & \bf 15\\ \cline{3-13} 
&&VN\_allo\_major & 90.62 & 89.57 & 88.48 & 66.76 & 57.80 & 62.26 & 91.49 & 90.73 & 90.51 & \bf 14\\ \cline{3-13} 
&&VN\_char\_mean & 91.91 & 91.24 & 90.47 & 68.27 & 59.13 & 64.17 & 92.53 & 92.07 & 92.29 & 10\\ \cline{3-13} 
&&VN\_allo\_mean & 92.74 & 90.91 & 91.29 & 69.23 & 60.04 & 64.37 & 93.43 & 91.89 & 92.63 & 9\\ \hline 
\parbox[t]{.01mm}{\multirow{4}{*}{\rotatebox[origin=c]{90}{\tiny{ResNet}}}}&
\parbox[t]{.01mm}{\multirow{4}{*}{\rotatebox[origin=c]{90}{\tiny{-101}}}}&RN\_char\_major & 92.01 & 92.43 & 91.76 & 69.27 & 59.66 & 65.24 & 93.62 & 92.97 & 92.86 & 7\\ \cline{3-13}
&&RN\_allo\_major & 92.44 & 91.72 & 91.05 & 69.69 & 59.61 & 64.98 & 93.42 & 92.33 & 92.27 & 8\\ \cline{3-13} 
&&RN\_char\_mean & 93.26 & 93.49 & 92.83 & 70.99 & 61.34 & 66.21 & 94.51 & 93.77 & 94.42 & 6\\ \cline{3-13} 
&&RN\_allo\_mean & 94.07 & 93.66 & 92.93 & 71.32 & 61.63 & 66.42 & 95.25 & 94.72 & 94.34 & 5\\ \hline 
\parbox[t]{.01mm}{\multirow{4}{*}{\rotatebox[origin=c]{90}{\tiny{Xception}}}}&
\parbox[t]{.01mm}{\multirow{4}{*}{\rotatebox[origin=c]{90}{\tiny{Net}}}}&XN\_char\_major & 95.56 & 94.53 & 93.73 & 71.47 & 62.51 & 67.54 & 95.51 & 95.73 & 94.82 & 3\\ \cline{3-13}
&&XN\_allo\_major & 94.83 & 93.66 & 93.50 & 70.93 & 62.58 & 66.83 & 95.07 & 95.19 & 94.48 & 4\\ \cline{3-13}
&&XN\_char\_mean & 96.73 & 95.46 & 95.12 & 72.75 & 63.74 & 68.66 & 97.04 & 96.97 & 96.12 & 2\\ \cline{3-13} 
&&XN\_allo\_mean & 97.02 & 95.71 & 95.47 & 73.74 & 64.12 & 68.94 & 97.87 & 96.46 & 96.84 & 1\\ \hline 
\end{tabular}\label{tab:table6}
\end{table}

%===================================================================================

\iffalse
X
.
Table 6. Top-1 writer identification performance using auto-derived features on $D_c$.

\fi

The ranks of the models are shown in the rightmost column of the Table \ref{tab:table6}. Here, the XN\_allo\_mean model performed best for writer identification on the $D_c$ database. 
Although in this model, the $AE_{ss}$, $AE_{smf/s}$ accuracies are more than 97\%, the performance is comparatively lower with respect to $AE_{smv}$, $AE_{sfv}$, $AE_{mfv}$ accuracies. 

In Table \ref{tab:table6}, the overall \emph{Strategy-Mean} worked better than \emph{Strategy-Major}. In general, the patch\textsubscript{allo} with the \emph{Strategy-Mean} worked better, but patch\textsubscript{allo} with \emph{Strategy-Major} did not work so well.

% 9.2.2.2. Writer Identification by Auto-derived Features on $D_{uc}$:
\paragraph{\ref{9.2.2}.2.  Writer Identification by Auto-derived Features on $D_{uc}$}~\\

In Table \ref{tab:table7}, we present the 9-tuple writer identification accuracies of auto-derived feature-based models performing on the $D_{uc}$ database. 

%===================================================================================

\begin{table}[!ht]
\tiny
\caption{Top-1 writer identification performance using auto-derived features on $D_{uc}$}
\centering
\begin{tabular}{rl|l|c|c|c|c|c|c|c|c|c|c}
\cline{3-13}
\multicolumn{1}{c}{}&\multicolumn{1}{c}{}&\multicolumn{1}{c|}{\textbf{Model}} & \multicolumn{9}{c|}{\textbf{9-tuple Accuracy (\%)}} & \multirow{2}{*}{\textbf{Rank}} \\ \cline{4-12} 
 \multicolumn{1}{c}{}&\multicolumn{1}{c}{}&\multicolumn{1}{c|}{($WI\_D_{uc}$)}& $AE_{ss}$ &  $AE_{mm}$ & $AE_{ff}$ & $AE_{smv}$ & $AE_{sfv}$ & $AE_{mfv}$ & $AE_{smf/s}$ & $AE_{smf/m}$ & $AE_{smf/f}$ & \\ \hline
\parbox[t]{0.01mm}{\multirow{4}{*}{\rotatebox[origin=c]{90}{\tiny{Basic}}}}&
 \parbox[t]{.01mm}{\multirow{4}{*}{\rotatebox[origin=c]{90}{\tiny{\_CNN}}}}& BCNN\_char\_major & 80.69 & 80.53 & 80.43 & 56.26 & 47.65 & 52.59 & 82.19 & 80.77 & 80.72 & 23\\ \cline{3-13}
&&BCNN\_allo\_major & 79.57 & 79.94 & 79.31 & 57.10 & 48.98 & 52.33 & 82.65 & 81.38 & 80.49 & 24\\ \cline{3-13} 
&&BCNN\_char\_mean & 81.96 & 82.05 & 82.78 & 58.16 & 49.47 & 53.04 & 84.22 & 83.23 & 82.01 & 22\\ \cline{3-13} 
&&BCNN\_allo\_mean & 83.01 & 81.73 & 81.64 & 59.5 & 50.02 & 54.15 & 85.13 & 84.24 & 82.81 & 21\\ \hline 
\parbox[t]{.01mm}{\multirow{4}{*}{\rotatebox[origin=c]{90}{\tiny{Squeeze}}}}&
\parbox[t]{.01mm}{\multirow{4}{*}{\rotatebox[origin=c]{90}{\tiny{Net}}}}&SN\_char\_major & 86.06 & 83.91 & 83.93 & 60.88 & 51.78 & 56.81 & 86.47 & 86.13 & 85.11 & 19\\ \cline{3-13} 
&&SN\_allo\_major & 85.25 & 83.51 & 83.29 & 62.26 & 50.96 & 55.71 & 86.13 & 85.43 & 83.85 & 20\\ \cline{3-13} 
&&SN\_char\_mean & 88.12 & 86.82 & 86.78 & 63.80 & 54.45 & 58.53 & 89.48 & 89.25 & 88.98 & 18\\ \cline{3-13} 
&&SN\_allo\_mean & 89.33 & 88.03 & 87.06 & 65.27 & 54.84 & 60.17 & 90.57 & 89.54 & 88.57 & 17\\ \hline 
\parbox[t]{.01mm}{\multirow{4}{*}{\rotatebox[origin=c]{90}{\tiny{GoogLe}}}}&
\parbox[t]{.01mm}{\multirow{4}{*}{\rotatebox[origin=c]{90}{\tiny{Net}}}}&GN\_char\_major & 90.31 & 89.96 & 87.82 & 66.14 & 55.75 & 60.79 & 90.85 & 91.34 & 89.82 & 13\\ \cline{3-13} 
&&GN\_allo\_major & 89.77 & 88.83 & 86.96 & 64.45 & 55.73 & 61.51 & 90.15 & 89.93 & 89.09 & 16\\ \cline{3-13} 
&&GN\_char\_mean & 90.43 & 90.47 & 88.39 & 66.41 & 56.48 & 60.86 & 92.14 & 90.79 & 91.07 & 12\\ \cline{3-13}
&&GN\_allo\_mean & 90.54 & 89.46 & 89.24 & 67.39 & 57.59 & 61.94 & 91.82 & 91.41 & 90.45 & 11\\ \hline 
\parbox[t]{.01mm}{\multirow{4}{*}{\rotatebox[origin=c]{90}{\tiny{VGG}}}}&
\parbox[t]{.01mm}{\multirow{4}{*}{\rotatebox[origin=c]{90}{\tiny{-16}}}}&VN\_char\_major & 90.37 & 89.55 & 87.51 & 65.67 & 55.82 & 60.54 & 91.49 & 90.69 & 90.52 & \bf 14\\ \cline{3-13} 
&&VN\_allo\_major & 89.74 & 88.85 & 88.63 & 66.92 & 55.64 & 60.41 & 91.61 & 89.20 & 90.41 & \bf 15\\ \cline{3-13} 
&&VN\_char\_mean & 91.60 & 90.18 & 90.38 & 67.09 & 57.39 & 63.12 & 90.78 & 90.49 & 91.65 & 10\\ \cline{3-13} 
&&VN\_allo\_mean & 92.69 & 90.34 & 90.55 & 67.94 & 58.46 & 63.02 & 92.52 & 91.55 & 91.46 & 9\\ \hline 
\parbox[t]{.01mm}{\multirow{4}{*}{\rotatebox[origin=c]{90}{\tiny{ResNet}}}}&
\parbox[t]{.01mm}{\multirow{4}{*}{\rotatebox[origin=c]{90}{\tiny{-101}}}}&RN\_char\_major & 91.82 & 92.72 & 90.28 & 68.16 & 58.95 & 63.42 & 93.16 & 92.96 & 92.35 & 7\\ \cline{3-13}
&&RN\_allo\_major & 91.85 & 91.42 & 90.73 & 68.60 & 57.41 & 63.07 & 92.42 & 91.41 & 91.97 & 8\\ \cline{3-13} 
&&RN\_char\_mean & 92.72 & 93.38 & 92.27 & 69.03 & 60.02 & 64.64 & 93.79 & 93.01 & 93.57 & 6\\ \cline{3-13} 
&&RN\_allo\_mean & 94.13 & 93.42 & 92.36 & 69.45 & 59.63 & 65.13 & 94.47 & 94.21 & 93.48 & 5\\ \hline 
\parbox[t]{.01mm}{\multirow{4}{*}{\rotatebox[origin=c]{90}{\tiny{Xception}}}}&
\parbox[t]{.01mm}{\multirow{4}{*}{\rotatebox[origin=c]{90}{\tiny{Net}}}}&XN\_char\_major & 95.13 & 94.31 & 92.82 & 69.15 & 60.89 & 65.77 & 95.27 & 95.34 & 93.85 & 3\\ \cline{3-13}
&&XN\_allo\_major & 94.62 & 93.01 & 92.51 & 69.95 & 61.51 & 65.78 & 94.42 & 94.84 & 94.16 & 4\\ \cline{3-13}
&&XN\_char\_mean & 96.49 & 94.56 & 94.56 & 70.85 & 61.85 & 67.81 & 96.25 & 96.04 & 95.74 & 2\\ \cline{3-13} 
&&XN\_allo\_mean & 96.81 & 95.52 & 95.03 & 72.52 & 62.79 & 66.53 & 97.09 & 96.59 & 95.62 & 1\\ \hline
\end{tabular}\label{tab:table7}
\end{table}

%===================================================================================

\iffalse
X
.
Table 7. Top-1 writer identification performance using auto-derived features on $D_{uc}$.

.
X
\fi

Here also, from Table \ref{tab:table7}, we note that performance on the intra-variable handwritten sample with different training and testing sample types is not so well, i.e., $AE_{smv}$, $AE_{sfv}$, $AE_{mfv}$ accuracies are comparatively low. The model XN\_allo\_mean produces the best outcome. The other model rankings can be observed in the rightmost column of Table \ref{tab:table7}. In this case, it can be noted that the overall \emph{Strategy-Mean} worked better than the \emph{Strategy-Major}. 

Comparing Tables \ref{tab:table6} and \ref{tab:table7}, it can be observed that the general performance on database $D_c$ is better than $D_{uc}$. The rank orders are almost similar in cases of Tables \ref{tab:table6} and \ref{tab:table7}. Here, only the ranks of model VN\_char\_major and VN\_allo\_major are interchanged. This scenario suggests that our model is quite stable for different databases.

All the auto-derived feature-based models worked better than handcrafted feature-based models. However, the $AE_{smv}$, $AE_{sfv}$, $AE_{mfv}$ accuracies are also low here in comparison with the other accuracies of the 9-tuple. Using auto-derived features, the Top-2 (Top-5) writer identification accuracies of $AE_{smv}$, $AE_{sfv}$, $AE_{mfv}$ increased at most 1.56\% (7.76\%) and 1.35\% (6.89\%) for $D_c$ and $D_{uc}$, respectively.

In Fig. \ref{fig:fig8} - Fig. \ref{fig:fig9}, we present the radar plots of 9-tuple accuracies of writer identification shown in 
Tables \ref{tab:table4} - \ref{tab:table7}.

% 9.3. Writer Verification Performance:===========================>
\subsection{Writer Verification Performance}
\label{sec.9.3}

In this section, we discuss writer verification performances of the models employed using both handcrafted features and auto-derived features.

The writer verification accuracies are obtained as mentioned in Section \ref{8WV}. Similar to writer identification, here we also use multiple experimental setups and finally obtain 9-tuple accuracies (refer to Section \ref{9.2.1}). 
The \emph{Borda count} \cite{51} is also used here to rank the models and shown in the last columns of Tables \ref{tab:table8} - \ref{tab:table11}.   

% 9.3.1. Writer Verification Performance with Handcrafted Features: 
\subsubsection{Writer Verification Performance with Handcrafted Features}
\label{9.3.1}

Similar to the three writer identification models employing handcrafted features (refer to Section \ref{9.2.1}), here also we obtain three writer verification models. These writer verification models are experimented on both the databases $D_c$ and $D_{uc}$. The procedure of model accuracy computation is described in Section \ref{sec.8.1}. The performance of the models is discussed as follows.

% 9.3.1.1. Writer Verification by Handcrafted Features on $D_c$:
\paragraph{\ref{9.3.1}.1. Writer Verification by Handcrafted Features on $D_c$}~\\

We have obtained $WV\_D_c\_F_{MM}$, $WV\_D_c\_F_{DH}$, $WV\_D_c\_F_{DC}$ models for writer verification similar to the writer identification models (refer to Section \ref{9.2.1}.1). 
For example, $WV\_D_c\_F_{MM}$ is such a writer verification model, where $F_{MM}$ handcrafted features are used on database $D_c$. 

In Table \ref{tab:table8}, we present the writer verification performance of these three models in terms of 9-tuple accuracy. The ranks of these models are mentioned in the rightmost column of Table \ref{tab:table8}. Here, the model $WV\_D_c\_F_{DH}$ performed best.

%===================================================================================

\begin{table}[!ht]
\tiny
\caption{Writer verification performance using handcrafted features on $D_{c}$}
\centering
\begin{tabular}{c|c|c|c|c|c|c|c|c|c|c}
\hline
\multirow{2}{*}{\textbf{Model}} & \multicolumn{9}{c|}{\textbf{9-tuple Accuracy (\%)}} & \multirow{2}{*}{\textbf{Rank}} \\ \cline{2-10} 
& $AE_{ss}$ &  $AE_{mm}$ & $AE_{ff}$ & $AE_{smv}$ & $AE_{sfv}$ & $AE_{mfv}$ & $AE_{smf/s}$ & $AE_{smf/m}$ & $AE_{smf/f}$ & \\ \hline
$WV\_D_{c}\_F_{MM}$ & 72.07 & 71.15 & 69.92 & 48.50 & 36.58 & 41.43 & 73.66 & 73.12 & 71.34 & 3\\ \hline
$WV\_D_{c}\_F_{DH}$ & 86.64 & 86.06 & 85.69 & 53.72 & 43.57 & 48.18 & 87.87 & 87.57 & 86.02 & 1\\ \hline
$WV\_D_{c}\_F_{DC}$ & 83.60 & 82.85 & 82.48 & 52.46 & 41.09 & 46.08 & 84.85 & 84.21 & 82.80 & 2\\ \hline
\end{tabular}\label{tab:table8}
\end{table}

%===================================================================================

\iffalse
x
.
Table 8. Writer verification performance using handcrafted features on $D_c$.

\fi

% 9.3.1.2. Writer Verification by Handcrafted Features on $D_{uc}$:
\paragraph{\ref{9.3.1}.2. Writer Verification by Handcrafted Features on $D_{uc}$}~\\

Here also, we have generated three handcrafted feature-based writer verification models $WV\_D_{uc}\_F_{MM}$, $WV\_D_{uc}\_F_{DH}$, $WV\_D_{uc}\_F_{DC}$ for experimentation on database $D_{uc}$.

In Table \ref{tab:table9}, the writer verification results of these three models are presented, where the rightmost column shows their ranking. In this case, the model $WV\_D_{uc}\_F_{DH}$ performed best.

%===================================================================================

\begin{table}[!ht]
\tiny
\caption{Writer verification performance using handcrafted features on $D_{uc}$}
\centering
\begin{tabular}{c|c|c|c|c|c|c|c|c|c|c}
\hline
\multirow{2}{*}{\bf Model} & \multicolumn{9}{c|}{\bf 9-tuple Accuracy (\%)} & \multirow{2}{*}{\bf Rank} \\ \cline{2-10} 
& $AE_{ss}$ &  $AE_{mm}$ & $AE_{ff}$ & $AE_{smv}$ & $AE_{sfv}$ & $AE_{mfv}$ & $AE_{smf/s}$ & $AE_{smf/m}$ & $AE_{smf/f}$ & \\ \hline
$WV\_D_{uc}\_F_{MM}$ & 69.58 & 69.05 & 67.09 & 47.96 & 35.76 & 40.26 & 71.43 & 70.12 & 69.34 & 3\\ \hline
$WV\_D_{uc}\_F_{DH}$ & 84.87 & 84.45 & 84.52 & 52.34 & 40.88 & 46.50 & 86.20 & 86.24 & 84.03 & 1\\ \hline
$WV\_D_{uc}\_F_{DC}$ & 81.50 & 80.38 & 81.12 & 50.86 & 39.32 & 45.74 & 83.09 & 83.05 & 81.34 & 2\\ \hline
\end{tabular}\label{tab:table9}
\end{table}

%===================================================================================

\iffalse
x
.
Table 9. Writer verification performance using handcrafted features on $D_{uc}$.

\fi

For experimentation on both databases $D_c$ and $D_{uc}$, the rankings are similar when using the same feature-based model. Overall, the $F_{DH}$ feature-based model performed best and the $F_{MM}$ feature-based model achieved lowest results for verification also.

It can be observed from Tables \ref{tab:table8} and \ref{tab:table9} that the accuracies $AE_{smv}$, $AE_{sfv}$, $AE_{mfv}$ are very low in comparison with other accuracies of 9-tuple. 
Here, the ranks in Tables \ref{tab:table8} and \ref{tab:table9} are the same for similar handcrafted feature-based models employed in $D_c$ and $D_{uc}$.

% 9.3.2. Writer Verification Performance with Auto-derived Features:
\subsubsection{Writer Verification Performance with Auto-derived Features}
\label{9.3.2}

In this section, we discuss the performance of auto-derived feature-based writer verification models. The procedure to obtain the verification accuracy using auto-derived features is mentioned in Section \ref{sec.8.2}. 
Here, only the \emph{Strategy-Mean} is used, since the \emph{Strategy-Major} performed insignificantly. Therefore, we have obtained 12 models from 6 types of convolutional networks, with 2 forms of patch (patch\textsubscript{char} and patch\textsubscript{allo}), fed using only one strategy. The naming convention of these verification models is kept similar to the writer identification models.

% 9.3.2.1. Writer Verification by Auto-derived Features on $D_c$:
\paragraph{\ref{9.3.2}.1. Writer Verification by Auto-derived Features on $D_c$}~\\

In Table \ref{tab:table10}, we present the 9-tuple writer verification accuracies using auto-derived features while experimenting on $D_c$. Here, XN\_allo\_mean performed best. All the model rankings are presented in the rightmost column of Table \ref{tab:table10}. 
In general, patch\textsubscript{allo} worked better than patch\textsubscript{char}. 
Only for VGG-16, the patch\textsubscript{char} performed better than patch\textsubscript{allo}.

%===================================================================================

\begin{table}[!ht]
\tiny
\caption{Writer verification performance using auto-derived features on  $D_{c}$}
\centering
% \rotatebox{90}{
\begin{tabular}{l|c|c|c|c|c|c|c|c|c|c}
\hline
\multicolumn{1}{c|}{\bf Model} & \multicolumn{9}{c|}{\bf 9-tuple Accuracy (\%)} & \multirow{2}{*}{\bf Rank} \\ \cline{2-10} 
\multicolumn{1}{c|}{($WV\_D_c$)}& $AE_{ss}$ &  $AE_{mm}$ & $AE_{ff}$ & $AE_{smv}$ & $AE_{sfv}$ & $AE_{mfv}$ & $AE_{smf/s}$ & $AE_{smf/m}$ & $AE_{smf/f}$ & \\ \hline
BCNN\_char\_mean & 92.22 & 91.48 & 92.13 & 69.19 & 60.94 & 66.59 & 94.17 & 93.53 & 93.44 & 12\\ \hline
BCNN\_allo\_mean & 92.45 & 91.78 & 92.75 & 70.69 & 61.84 & 66.28 & 93.33 & 92.12 & 92.72 & 11\\ \hline
SN\_char\_mean & 95.74 & 94.85 & 95.53 & 74.85 & 65.14 & 70.97 & 97.24 & 97.31 & 96.55 & 10\\ \hline
SN\_allo\_mean & 96.09 & 95.09 & 96.52 & 77.08 & 66.16 & 70.75 & 96.53 & 95.93 & 96.07 & 9\\ \hline
GN\_char\_mean & 96.55 & 95.60 & 96.21 & 75.39 & 65.97 & 71.42 & 97.97 & 97.97 & 96.73 & \bf 8\\ \hline
GN\_allo\_mean & 96.24 & 95.83 & 96.55 & 77.90 & 66.18 & 71.47 & 97.38 & 96.79 & 96.58 & \bf 7\\ \hline
VN\_char\_mean & 97.77 & 96.36 & 97.02 & 75.80 & 66.69 & 72.19 & 98.39 & 98.17 & 97.67 & \bf 5\\ \hline
VN\_allo\_mean & 97.20 & 96.81 & 96.73 & 78.04 & 66.48 & 72.35 & 97.82 & 96.96 & 97.08 & \bf 6\\ \hline
RN\_char\_mean & 98.21 & 97.85 & 98.24 & 77.48 & 67.83 & 73.28 & 99.50 & 99.47 & 99.05 & 4\\ \hline
RN\_allo\_mean & 98.74 & 98.15 & 98.27 & 79.72 & 68.36 & 73.51 & 99.76 & 98.73 & 98.95 & 3\\ \hline
XN\_char\_mean & 98.95 & 98.79 & 98.82 & 78.68 & 69.70 & 75.24 & 99.76 & 99.68 & 99.27 & 2\\ \hline
XN\_allo\_mean & 99.24 & 99.06 & 98.68 & 80.79 & 70.02 & 74.98 & 99.84 & 99.73 & 99.18 & 1\\ \hline
\end{tabular}\label{tab:table10}
% }
\end{table}

%===================================================================================

\iffalse
X
.
Table 10. Writer verification performance using auto-derived features on $D_c$.

\fi

% 9.3.2.2. Writer Verification by Auto-derived Features on $D_{uc}$:
\paragraph{\ref{9.3.2}.2. Writer Verification by Auto-derived Features on $D_{uc}$}~\\

On database $D_{uc}$, the 9-tuple writer verification accuracies of various auto-derived feature-based models are shown in Table \ref{tab:table11}. In this case, XN\_allo\_mean performed best. The ranks of other models are shown in the rightmost column of Table \ref{tab:table11}. 
Overall, patch\textsubscript{allo} worked better than patch\textsubscript{char} except for VGG-16 and GoogLeNet.

%===================================================================================

\begin{table}[!ht]
\tiny
\caption{Writer verification performance using auto-derived features on  $D_{uc}$}
\centering
% \rotatebox{90}{
\begin{tabular}{l|c|c|c|c|c|c|c|c|c|c}
\hline
\multicolumn{1}{c|}{\bf Model} & \multicolumn{9}{c|}{\bf 9-tuple Accuracy (\%)} & \multirow{2}{*}{\bf Rank} \\ \cline{2-10} 
\multicolumn{1}{c|}{($WV\_D_{uc}$)}& $AE_{ss}$ &  $AE_{mm}$ & $AE_{ff}$ & $AE_{smv}$ & $AE_{sfv}$ & $AE_{mfv}$ & $AE_{smf/s}$ & $AE_{smf/m}$ & $AE_{smf/f}$ & \\ \hline
BCNN\_char\_mean & 91.55 & 90.57 & 91.26 & 69.15 & 60.17 & 65.54 & 93.24 & 92.57 & 92.06 & 12\\ \hline
BCNN\_allo\_mean & 91.78 & 90.92 & 91.78 & 70.29 & 61.31 & 65.91 & 92.63 & 91.81 & 92.53 & 11\\ \hline
SN\_char\_mean & 95.36 & 94.25 & 94.54 & 74.77 & 64.15 & 69.98 & 96.50 & 96.95 & 96.12 & 10\\ \hline
SN\_allo\_mean & 95.86 & 94.46 & 96.50 & 76.84 & 65.35 & 70.54 & 96.06 & 95.90 & 95.47 & 9\\ \hline
GN\_char\_mean & 95.86 & 95.29 & 96.02 & 74.95 & 65.36 & 71.27 & 97.07 & 97.09 & 96.67 & \bf 7\\ \hline
GN\_allo\_mean & 95.55 & 95.29 & 96.54 & 77.14 & 65.81 & 70.52 & 97.23 & 96.76 & 96.12 & \bf 8\\ \hline
VN\_char\_mean & 97.36 & 96.02 & 96.72 & 75.10 & 66.58 & 71.62 & 97.06 & 97.94 & 97.62 & \bf 5\\ \hline
VN\_allo\_mean & 96.43 & 96.75 & 95.99 & 77.28 & 66.18 & 72.34 & 97.67 & 96.61 & 96.45 & \bf 6\\ \hline
RN\_char\_mean & 97.46 & 97.63 & 98.02 & 77.15 & 67.22 & 73.03 & 99.30 & 99.29 & 98.53 & 4\\ \hline
RN\_allo\_mean & 98.44 & 98.03 & 98.27 & 79.40 & 68.26 & 73.37 & 98.89 & 97.79 & 98.13 & 3\\ \hline
XN\_char\_mean & 98.73 & 98.69 & 98.06 & 78.13 & 69.29 & 74.28 & 99.65 & 98.96 & 98.77 & 2\\ \hline
XN\_allo\_mean & 98.55 & 98.73 & 97.89 & 79.84 & 69.80 & 74.76 & 99.19 & 99.54 & 98.54 & 1\\ \hline
\end{tabular}\label{tab:table11}
% }
\end{table}

%===================================================================================

\iffalse
X
.
Table 11. Writer verification performance using auto-derived features on $D_{uc}$.

\fi

From Tables \ref{tab:table10} and \ref{tab:table11}, it can be observed that the $AE_{smv}$, $AE_{sfv}$, $AE_{mfv}$ accuracies are very low in comparison with other accuracies of 9-tuple. Here, Tables \ref{tab:table10} and \ref{tab:table11} depict similar ranks except for the GoogLeNet-based models.

In Fig. \ref{fig:fig10} - Fig. \ref{fig:fig11}, we present the radar plots of 9-tuple accuracies of writer verification shown in 
Tables \ref{tab:table8} - \ref{tab:table11}.

% 9.4. Observations ======================================>
\subsection{Observations}

\begin{figure}[!b]
  \centering
   \includegraphics[width=\linewidth]{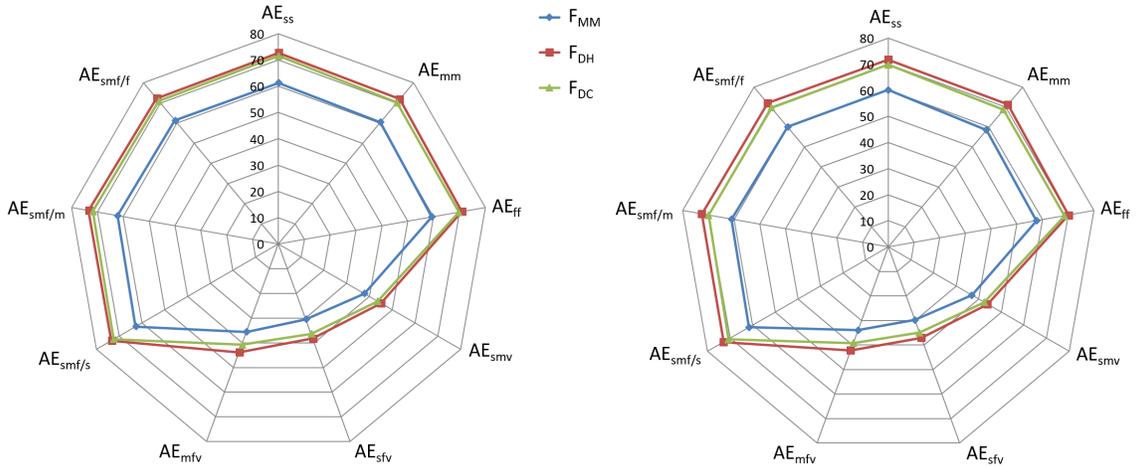}
  \caption{Radar plot of Top-1 writer identification performance using handcrafted features. \emph{Left}: representing Table \ref{tab:table4} on $D_c$ and \emph{Right}: representing Table \ref{tab:table5} on $D_{uc}$.}
  \label{fig:fig8}
 \end{figure}

\begin{figure}[!t]
  \centering
   \includegraphics[width=\linewidth]{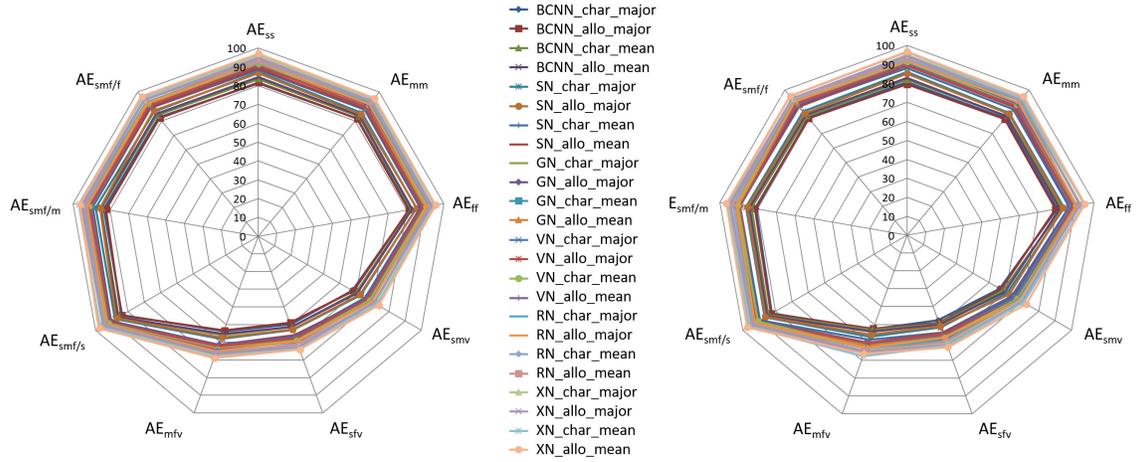}
  \caption{Radar plot of Top-1 writer identification performance using auto-derived features. \emph{Left}: representing Table \ref{tab:table6} on $D_c$ and \emph{Right}: representing Table \ref{tab:table7} on $D_{uc}$.}
  \label{fig:fig9}
 \end{figure}

\begin{figure}[!t]
  \centering
   \includegraphics[width=\linewidth]{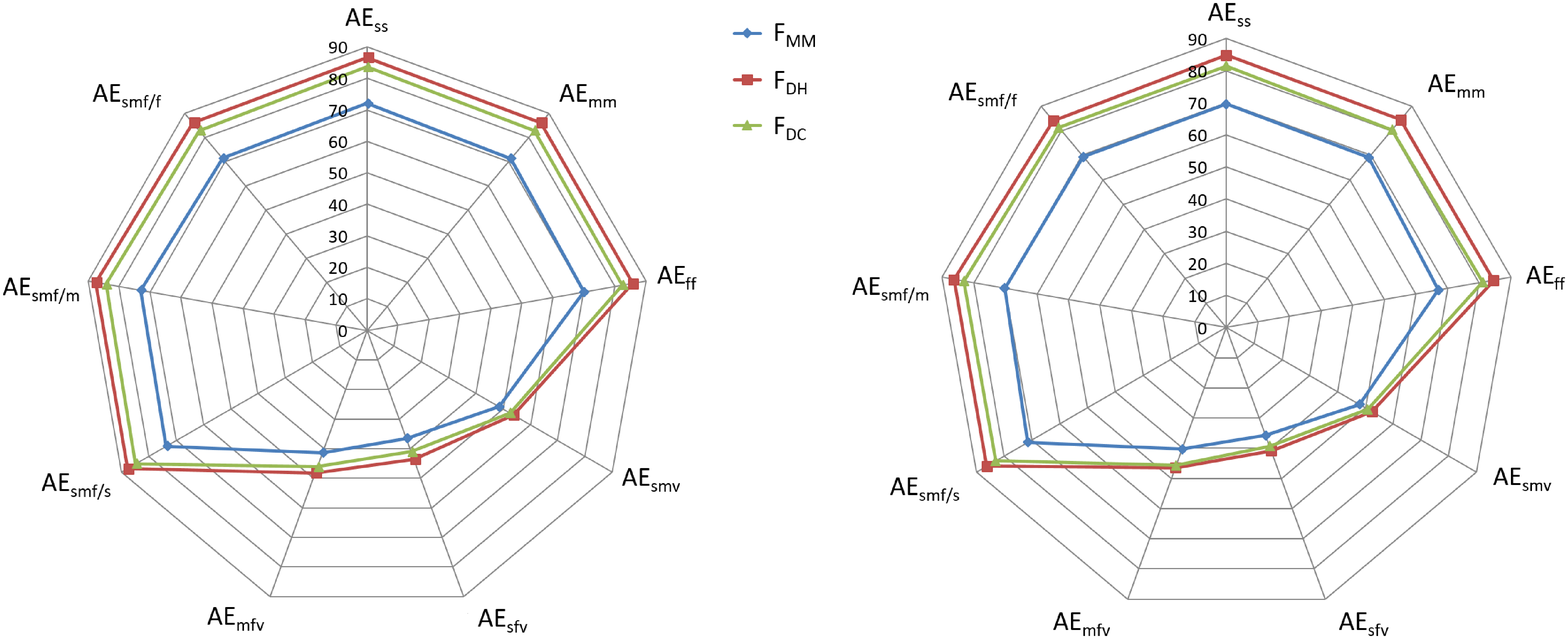}
  \caption{Radar plot of writer verification performance using handcrafted features. \emph{Left}: representing Table \ref{tab:table8} on $D_c$ and \emph{Right}: representing Table \ref{tab:table9} on $D_{uc}$.}
  \label{fig:fig10}
 \end{figure}

\begin{figure}[!t]
  \centering
   \includegraphics[width=\linewidth]{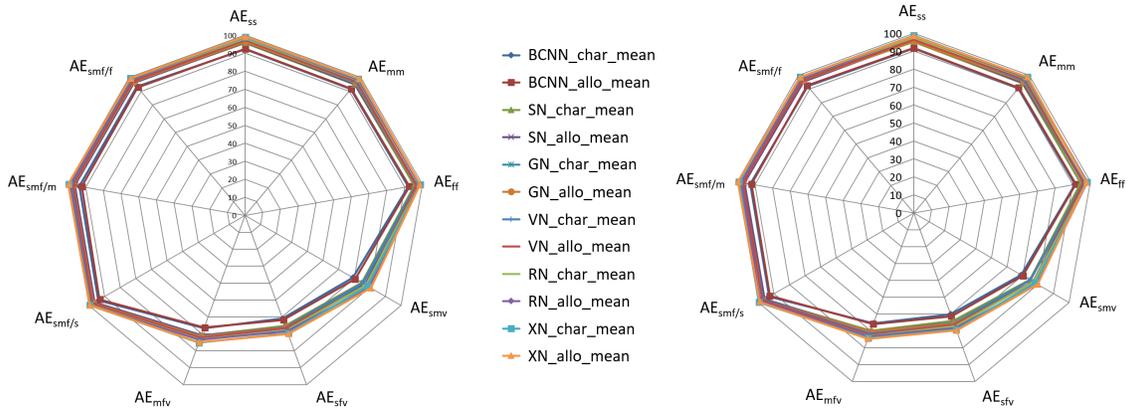}
  \caption{Radar plot of writer verification performance using auto-derived features. \emph{Left}: representing Table \ref{tab:table10} on $D_c$ and \emph{Right}: representing Table \ref{tab:table11} on $D_{uc}$.}
  \label{fig:fig11}
 \end{figure}

From the above experiments (refer to Sections \ref{sec.9.2} and \ref{sec.9.3}, Tables \ref{tab:table4} - \ref{tab:table11}, Fig. \ref{fig:fig8} - Fig. \ref{fig:fig11}), our major observations are noted as follows:

i) Among the accuracies in 9-tuple, the $AE_{smv}$, $AE_{sfv}$, $AE_{mfv}$ accuracies are comparatively low for all the models used for \emph{writer identification/verification}. 
It suggests that if the training and test set contain similar types of samples, the models can perform better.
% It suggests that if the training set contains similar types of test set samples, the models can perform better.

Moreover, here the $AE_{sfv}$ accuracy is lower than $AE_{smv}$ and $AE_{mfv}$. It indicates that sets $S_s$ and $S_f$ of $D_c$ ($S'_s$ and $S'_f$ sets of $D_{uc}$) contain higher intra-variable writing than the other set combinations.

ii) The auto-derived features worked better than the handcrafted features for the \emph{writer identification/verification}. The reason may be the high dimensionality of the auto-derived features and the use of deep convolutional architectures.

Among the handcrafted features, $F_{DH}$ performed best, while $F_{DC}$ performed better than $F_{MM}$. Among auto-derived features, Xception Net-based features worked best.

iii) For auto-derived feature-based \emph{writer identification} models, mostly the \emph{Strategy-Mean} worked better than \emph{Strategy-Major}.

iv) In general, for auto-derived feature-based \emph{writer identification} models, patch\textsubscript{allo} with \emph{Strategy-Mean} worked best, whereas patch\textsubscript{allo} with \emph{Strategy-Major} performed lowest. Combining patch\textsubscript{char}, \emph{Strategy-Mean} worked better than \emph{Strategy-Major}. 
In other words, using combinations of \emph{patch} and \emph{Strategy}, the overall performance in highest to lowest order is as follows: 
allo\_mean $\succ$ char\_mean $\succ$ char\_major $\succ$ allo\_major.

v) For auto-derived feature-based \emph{writer verification} models, mostly the patch\textsubscript{allo} worked better than the patch\textsubscript{char}.

vi) All together, the \emph{writer identification/verification} performance on the controlled database ($D_c$) is better than the uncontrolled database ($D_{uc}$).

vii) As a whole, all the \emph{writer identification/verification} models provide quite similar $AE_{ss}$, $AE_{mm}$, $AE_{ff}$ accuracies (differences range up to 3.12\% Top-1). This implies that our system is quite robust on working with various handwriting types. 

viii) In general, the handcrafted feature-based models and auto-derived feature-based models for \emph{writer identification/verification} follow the same trend, respectively (refer to Tables \ref{tab:table4} - \ref{tab:table11}). It can be visualized by the radar plots of Fig. \ref{fig:fig8} - Fig. \ref{fig:fig11}. Here, the individual radar plot follows almost the same trend with creating a band of certain width.

% 9.5. Writer Identification/Verification by Pre-training ===========================>
\subsection{Writer Identification/Verification by Pre-training}
\label{sec.9.5}

From the previous experiments, we have observed that $AE_{smv}$, $AE_{sfv}$, $AE_{mfv}$ accuracies are relatively lower than other accuracies of the 9-tuple. Therefore, in this section, we aim to increase these three accuracies ($AE_{smv}$, $AE_{sfv}$, $AE_{mfv}$), say, by 3-tuple.

We have noted that the auto-derived features worked better than the handcrafted features. Therefore, here we focus only on auto-derived features. We have also observed that the \emph{Xception Net} performed best among all models for our problem. Hence, we continue investigations with this network for increasing the 3-tuple ($AE_{smv}$, $AE_{sfv}$, $AE_{mfv}$) accuracies, and present them here and further in this paper.

It can be reiterated that the 100 writers contributing to the $D_c$ database are completely different from the 100 writers in $D_{uc}$. As a matter of fact, there is no writer overlap between $D_c$ and $D_{uc}$ (refer to Section \ref{sec.3.2.Duc}).

We pre-train the model using $D_{uc}$ and repeat our experiments on $D_c$ (say, $E\_D_{uc}/D_c$). We get this idea from the \emph{transfer learning} approach \cite{69}. For pre-training, the total database $D_{uc}$ is used. Then previous experimental setups, i.e., $E_{sm}$, $E_{ms}$, $E_{sf}$, $E_{fs}$, $E_{mf}$ and $E_{fm}$, are used to obtain $AE_{smv}$, $AE_{sfv}$, $AE_{mfv}$ accuracies (refer to Section \ref{sec.9.2}). For example, we have pre-trained the model using all the data of $D_{uc}$, now for the $E_{sm}$ setup, we again train the model with $S_{s1}$ and test on $S_{m3}$.  

%===================================================================================

\begin{table}[!b]
\tiny
\caption{Top-1 writer identification by pre-training}
\centering
\begin{tabular}{c|c|c|c|c V{3}c|c|c|c|c}
\hline
\multirow{2}{*}{\bf Setup} & \bf{Model} & \multicolumn{3}{c V{3}}{\bf 3-tuple Accuracy (\%)} & \multirow{2}{*}{\bf Setup} & \bf{Model} & \multicolumn{3}{c}{\bf 3-tuple Accuracy (\%)}
\\ \cline{3-5}\cline{8-10}
&($WI$)& $AE_{smv}$ & $AE_{sfv}$ & $AE_{mfv}$ &&($WI$)& $AE_{smv}$ & $AE_{sfv}$ & $AE_{mfv}$
\\ \hline 
\multirow{4}{*}{$E\_D_{uc}/D_{c}$} & XN\_char\_major & 74.68 & 63.92 & \bf 68.09 & \multirow{4}{*}{$E\_D_{c}/D_{uc}$} & XN\_char\_major & 74.03 & 63.89 & 67.75
\\ \cline{2-5}\cline{7-10}
& XN\_allo\_major & 73.64 & 63.38 & \bf 68.47 & & XN\_allo\_major & 72.78 & 63.24 & 67.54
\\\cline{2-5}\cline{7-10}
& XN\_char\_mean & 76.31 & 66.38 & 71.03 & & XN\_char\_mean & 75.70 & 65.41 & 70.87
\\\cline{2-5}\cline{7-10}
& XN\_allo\_mean & 77.37 & 67.21 & 71.95 & & XN\_allo\_mean & 76.57 & 66.56 & 71.76
\\\hline
% \\\cline{2-5}
% \\\cline{2-5}
% \\\cline{2-5}
% \\ \hline 
\end{tabular}\label{tab:table12}
\end{table}

%===================================================================================

Likewise, the $E\_D_c/D_{uc}$ experiment is performed where the total $D_c$ is used for pre-training, and then the earlier experiments are repeated on $D_{uc}$ (refer to Section \ref{sec.9.2}, \ref{sec.9.3}). Here, two databases have assisted each other in pre-training/learning to study the system performance. Such a technique may be referred to as \emph{cross-learning}.

In Table \ref{tab:table12}, we present the Top-1 \emph{writer identification} performances on $E\_D_{uc}/D_c$ and $E\_D_c/D_{uc}$ experimental setups employing various \emph{Xception Net} models.

From Table \ref{tab:table12}, it can be observed that substantially the XN\_allo\_mean performed best for both $E\_D_{uc}/D_c$ and $E\_D_c/D_{uc}$ setups. Overall, on 3-tuple accuracy, the highest-to-lowest performance order is as follows: 
XN\_allo\_mean $\succ$ XN\_char\_mean $\succ$ XN\_char\_major $\succ$ XN\_allo\_major. 
The only exception is to obtain $AE_{mfv}$ accuracy in $E\_D_{uc}/D_c$, where the XN\_allo\_major worked better than XN\_char\_major.

\iffalse
%===================================================================================

\begin{table}[!ht]
\scriptsize
\caption{Top-1 writer identification by pre-training}
\centering
\begin{tabular}{l|c|c|c|c}
\hline
\multirow{2}{*}{Setup} & Model & \multicolumn{3}{c}{3-tuple Accuracy (\%)}\\ \cline{3-5}
&(WI)& $AE_{smv}$ & $AE_{sfv}$ & $AE_{mfv}$\\ \hline 
\multirow{4}{*}{$E\_D_{uc}/D_{c}$} & XN\_char\_major & 74.68 & 63.92 & 68.09\\ \cline{2-5}
& XN\_allo\_major & 73.64 & 63.38 & 68.47\\\cline{2-5}
& XN\_char\_mean & 76.31 & 66.38 & 71.03\\\cline{2-5}
& XN\_allo\_mean & 77.37 & 67.21 & 71.95\\\hline
\multirow{4}{*}{$E\_D_{c}/D_{uc}$} & XN\_char\_major & 74.03 & 63.89 & 67.75\\\cline{2-5}
& XN\_allo\_major & 72.78 & 63.24 & 67.54\\\cline{2-5}
& XN\_char\_mean & 75.7 & 65.41 & 70.87\\\cline{2-5}
& XN\_allo\_mean & 76.57 & 66.56 & 71.76\\ \hline 
\end{tabular}\label{tab:table12}
\end{table}

%===================================================================================
\fi
\iffalse
Table 12. Top-1 writer identification by pre-training

\fi

Overall, for $E\_D_{uc}/D_c$ and $E\_D_c/D_{uc}$ experiments, the Top-2 (Top-5) writer identification criteria produced up to 1.86\% (7.52\%) and 2.08\% (8.37\%) additional accuracy, respectively.

The \emph{writer verification} performance for the experimental setups $E\_D_{uc}/D_c$ and $E\_D_c/D_{uc}$ are presented in Table \ref{tab:table13}. 
Here, for both the $E\_D_{uc}/D_c$ and $E\_D_c/D_{uc}$ experiments, mostly patch\textsubscript{allo} worked better than patch\textsubscript{char}. 
The only exception is to obtain $AE_{smv}$ in $E\_D_c/D_{uc}$ where XN\_char\_mean worked better than XN\_allo\_mean.

%===================================================================================

\begin{table}[!ht]
%\scriptsize
\tiny
\caption{Writer verification by pre-training}
\centering
\begin{tabular}{c|c|c|c|c}
\hline
\multirow{2}{*}{\bf Setup} & \bf{Model} & \multicolumn{3}{c}{\bf 3-tuple Accuracy (\%)}\\ \cline{3-5}
&($WV$)& $AE_{smv}$ & $AE_{sfv}$ & $AE_{mfv}$\\ \hline 
\multirow{2}{*}{$E\_D_{uc}/D_{c}$} & XN\_char\_mean & 83.05 & 72.21 & 76.82\\ \cline{2-5}
& XN\_allo\_mean & 83.54 & 72.45 & 77.65\\ \hlineB{2.5} %\hline
\multirow{2}{*}{$E\_D_{c}/D_{uc}$} & XN\_char\_mean & \bf 82.58 & 71.69 & 76.25\\\cline{2-5}
& XN\_allo\_mean & \bf 82.34 & 72.47 & 77.76\\ \hline 
\end{tabular}\label{tab:table13}
\end{table}

%===================================================================================

\iffalse
Table 13.  Writer verification by pre-training

\fi

Compared to the results of Section \ref{sec.9.2} and \ref{sec.9.3}, here it can be seen that such pre-training with cross-learning has improved the system performance (at most 5.23\% for identification and 3.00\% for verification).

Now, comparing the experimental setups of Tables \ref{tab:table12} and \ref{tab:table13}, we observe that $E\_D_{uc}/D_c$ performed better than $E\_D_c/D_{uc}$ for both identification and verification. Here also, all the models/experimental setups are compared using the \emph{Borda count} \cite{51}.

% 9.6. Writer Identification/Verification on the Enlarged Writer Set =======================>
\subsection{Writer Identification/Verification on the Enlarged Writer Set}

In this section, we would like to see the system performance on an increased number of writers. For this purpose, we merged the controlled ($D_c$) and uncontrolled ($D_{uc}$) databases and get data from 200 writers. This experimental setup is denoted as $E\_D_{c+uc}$.

Here also, we show the 3-tuple accuracies of \emph{Xception Net}-based models as in Section \ref{sec.9.5}, although we have computed 9-tuple accuracies from all models (refer to Section \ref{sec.9.2}, \ref{sec.9.3}).

The earlier experimental setups, i.e., $E_{sm}$, $E_{ms}$, $E_{sf}$, $E_{fs}$, $E_{mf}$ and $E_{fm}$ setups, are used here to obtain $AE_{smv}$, $AE_{sfv}$, $AE_{mfv}$ accuracies (refer to Section \ref{sec.9.2}). For example, the setup $E_{sm}$ of $E\_D_{c+uc}$ is trained by the $S_{s1}+ S'_{s1}$ set and is tested on the $S_{m3}+ S'_{m3}$ set. 

In Table \ref{tab:table14}, we present the Top-1 \emph{writer identification} performance in terms of 3-tuple accuracy, which shows that XN\_allo\_mean performed best. Overall, on 3-tuple accuracy, the performance from highest to lowest order is as follows: 
XN\_allo\_mean $\succ$ XN\_char\_mean $\succ$ XN\_char\_major $\succ$ XN\_allo\_major. 
However, there is an exception for $AE_{sfv}$ where XN\_allo\_major has worked better than XN\_char\_major.

%===================================================================================
\begin{table}[!ht]
\tiny %\scriptsize
\caption{Top-1 writer identification on enlarged writer set}
\centering
\begin{tabular}{c|c|c|c}
\hline
\bf {Model} & \multicolumn{3}{c}{\bf 3-tuple Accuracy (\%)}\\ \cline{2-4}
($WI\_E\_D_{c+uc}$)& $AE_{smv}$ & $AE_{sfv}$ & $AE_{mfv}$\\ \hline 
XN\_char\_major & 73.96 & \bf 63.84 & 67.27\\ \hline
XN\_allo\_major & 72.72 & \bf 64.02 & 67.16\\\hline
XN\_char\_mean & 76.07 & 66.37 & 70.79\\\hline
XN\_allo\_mean & 77.76 & 66.72 & 70.95\\ \hline 
\end{tabular}\label{tab:table14}
\end{table}
%===================================================================================

\iffalse
Table 14. Top-1  writer identification on enlarged writer set

\fi

Overall for $E\_D_{c+uc}$, the Top-2 and Top-5 writer identification criteria produced up to 0.47\% and 4.38\% additional accuracies, respectively.

In Table \ref{tab:table15}, we present the \emph{writer verification} performance of the $E\_D_{c+uc}$ setup. In this case, XN\_allo\_mean worked better than XN\_char\_mean.

%===================================================================================

\begin{table}[!ht]
\tiny %\scriptsize
\caption{Top-1 writer verification on an enlarged writer set}
\centering
\begin{tabular}{c|c|c|c}
\hline
\bf{Model} & \multicolumn{3}{c}{\bf 3-tuple Accuracy (\%)}\\ \cline{2-4}
($WV\_E\_D_{c+uc}$)& $AE_{smv}$ & $AE_{sfv}$ & $AE_{mfv}$\\ \hline 
XN\_char\_mean & 83.68 & 72.65 & 77.19\\ \hline
XN\_allo\_mean & 84.02 & 72.81 & 77.87\\\hline
\end{tabular}\label{tab:table15}
\end{table}
%===================================================================================

\iffalse
Table 15. Top-1  writer verification on an enlarged writer set

\fi

For writer identification/verification on $E\_D_{c+uc}$, XN\_allo\_mean performed best. Here, all the models are compared using \emph{Borda count} \cite{51}.

By comparing with the results from Sections \ref{sec.9.2} and \ref{sec.9.3}, we see that the system performance has slightly increased (at most 4.02\% for identification and 3.23\% for verification) with the number of writers.

% 9.7. Comparison with Other Works: ======================>
\subsection{Comparison with Other Works}

To the best of our knowledge and understanding, our work is the earliest attempt of its kind on such a problem. Also, we did not find any other published research work on this topic to be compared.

%% file: 10conclusion.tex
\section{Conclusion}
\label{10conclusion}

%10. Conclusion

In this paper, we work on writer identification/verification when there is extensive variation in a person's handwriting. 
In brief, we focus on high intra-variable handwriting-based writer investigation. 
We employ both handcrafted and auto-derived feature-based models to study writer identification/verification performance. We generated two offline Bengali intra-variable handwriting databases from two different sets of 100 writers. For this database generation, we have also worked with auto-derived feature-based grouping technique to form similar groups of intra-variable writing. After experimenting on our databases, we observe that by training and testing on similar writing variability, our system produces encouraging outcomes. However, our system performance is comparatively lower for training and testing on disparate types of handwriting variability. We also attempt with cross-learning and see that the system performance improves with pre-training.   

Here, a practical scenario is imitated, whereby a certain writing style of an individual is unknown (i.e., absent during training), 
and we note that the state-of-the-art methods do not perform 
% equally well similar to the ideal situation. 
well. 
However, we also observe that the deep features have high potential for this task. In future, we will try to exploit this potential and find some latent characteristics of a person from his/her varying styles of writing.

\section*{Acknowledgment}
 We heartily thank all the volunteers for their immense help in generating our database. 
 %The advice from the Centre for Forensic Science, UTS is also gratefully acknowledged.
%We thank all the volunteers for their help in generating our database. The advice from the Centre for Forensic Science, UTS is gratefully acknowledged.